\documentclass{article}

\usepackage{PRIMEarxiv}

\usepackage[utf8]{inputenc} 
\usepackage[T1]{fontenc}    
\usepackage{hyperref}       
\usepackage{url}            
\usepackage{booktabs}       
\usepackage{amsfonts}       
\usepackage{nicefrac}       
\usepackage{microtype}      
\usepackage{lipsum}
\usepackage{fancyhdr}       
\usepackage{graphicx}       
\graphicspath{{media/}}     

\usepackage{microtype}
\usepackage{graphicx}
\usepackage{subfigure}
\usepackage{booktabs} 
\usepackage{makecell} 
\usepackage{booktabs} 

\usepackage{microtype}
\usepackage{graphicx}
\usepackage{subfigure}
\usepackage{booktabs} 
\usepackage{makecell} 
\usepackage{booktabs} 

\usepackage{hyperref}

\usepackage{amsmath}
\usepackage{amssymb}
\usepackage{mathtools}
\usepackage{amsthm}

\usepackage{algorithm}
\usepackage{algorithmic}
\usepackage{scalefnt}

\title{Psychometric-Based Evaluation for Theorem Proving with Large Language Models
}

\author{
	Jianyu Zhang\textsuperscript{1}, Yongwang Zhao\textsuperscript{2,*}, Long Zhang\textsuperscript{3}, Jilin Hu\textsuperscript{2}, Xiaokun Luan\textsuperscript{4}, Zhiwei Xu\textsuperscript{2}, Feng Yang\textsuperscript{3}\\
	\textsuperscript{1}University of Electronic Science and Technology of China, \texttt{202222060304@std.uestc.edu.cn} \\
	\textsuperscript{2}Zhejiang University \\
	\textsuperscript{3}Academy of Military Sciences \\
	\textsuperscript{4}Peking University \\
	\texttt{*Corresponding author: zhaoyw@zju.edu.cn}
}

\begin{document}
\maketitle

\begin{abstract}
Large language models (LLMs) for formal theorem proving have become a prominent research focus. At present, the proving ability of these LLMs is mainly evaluated through proof pass rates on datasets such as miniF2F. However, this evaluation method overlooks the varying importance of theorems. As a result, it fails to highlight the real performance disparities between LLMs and leads to high evaluation costs. This study proposes a psychometric-based evaluation method for theorem proving with LLMs, comprising two main components: Dataset Annotation and Adaptive Evaluation. First, we propose a metric calculation method to annotate the dataset with difficulty and discrimination metrics. Specifically, we annotate each theorem in the miniF2F dataset and grade them into varying difficulty levels according to the performance of LLMs, resulting in an enhanced dataset: miniF2F-Graded. Experimental results show that the difficulty grading in miniF2F-Graded better reflects the theorem difficulty perceived by LLMs. Secondly, we design an adaptive evaluation method to dynamically select the most suitable theorems for testing based on the annotated metrics and the real-time performance of LLMs. We apply this method to evaluate 10 LLMs. The results show that our method finely highlights the performance disparities between LLMs. It also reduces evaluation costs by using only 23\% of the theorems in the dataset.
\end{abstract}


\section{Introduction}\label{section-1}
The rapid development of large language models (LLMs) has led to the creation of specialized models across various fields \cite{naveed2023comprehensive, zhao2023survey}. One area that has garnered significant attention is theorem proving in formal methods. The application of theorem proving is widespread, especially in mathematical research and in verifying the correctness and security of safety-critical systems. However, theorem proving remains a complex and labor-intensive task, often requiring substantial effort from experts to complete \cite{yang2024formal, ahn2024large}. The advent of LLMs has made automated theorem proving feasible. Pre-trained LLMs can now assist developers in rapidly completing complex proof tasks, such as DeepSeek-Prover \cite{xin2024deepseekproverv15harnessingproofassistant} and LEGO-Prover \cite{wang2023lego}.

In automated theorem proving, evaluating the theorem-proving ability of LLMs is a crucial task, as accurate evaluation standards are essential to guide the continuous improvement of LLM training methods \cite{xia2024evaluating, yang2024formal}. Currently, the mainstream evaluation method is testing the proportion of theorems successfully proven by the model within N attempts ($Pass@N$). While this evaluation method is intuitive, it faces several challenges:

\begin{itemize} 
	\item The evaluation results fail to highlight the disparities in theorem-proving abilities between LLMs, as the pass rate overlooks the varying importance of theorems and assigns equal weight to both difficult and simple ones. For instance, DeepSeek-prover-RL introduced several meaningful improvements over DeepSeek-Prover-SFT but only achieved a slight increase in $Pass@128$ \cite{xin2024deepseekproverv15harnessingproofassistant}, suggesting that the evaluation results lack a fine-grained distinction.
	
	\item Evaluation incurs high computational costs, as it does not filter the test theorems and requires testing all the theorems in the dataset. For example, testing the $Pass@128$ score of a 7B-metric model on 488 theorems in our experiment took 39.3 hours on a single A100 GPU. For larger models or datasets, the computational cost would increase even further.
\end{itemize}

\begin{figure*}[ht]
	\vskip 0.2in
	\begin{center}
		\centerline{\includegraphics[width=\columnwidth]{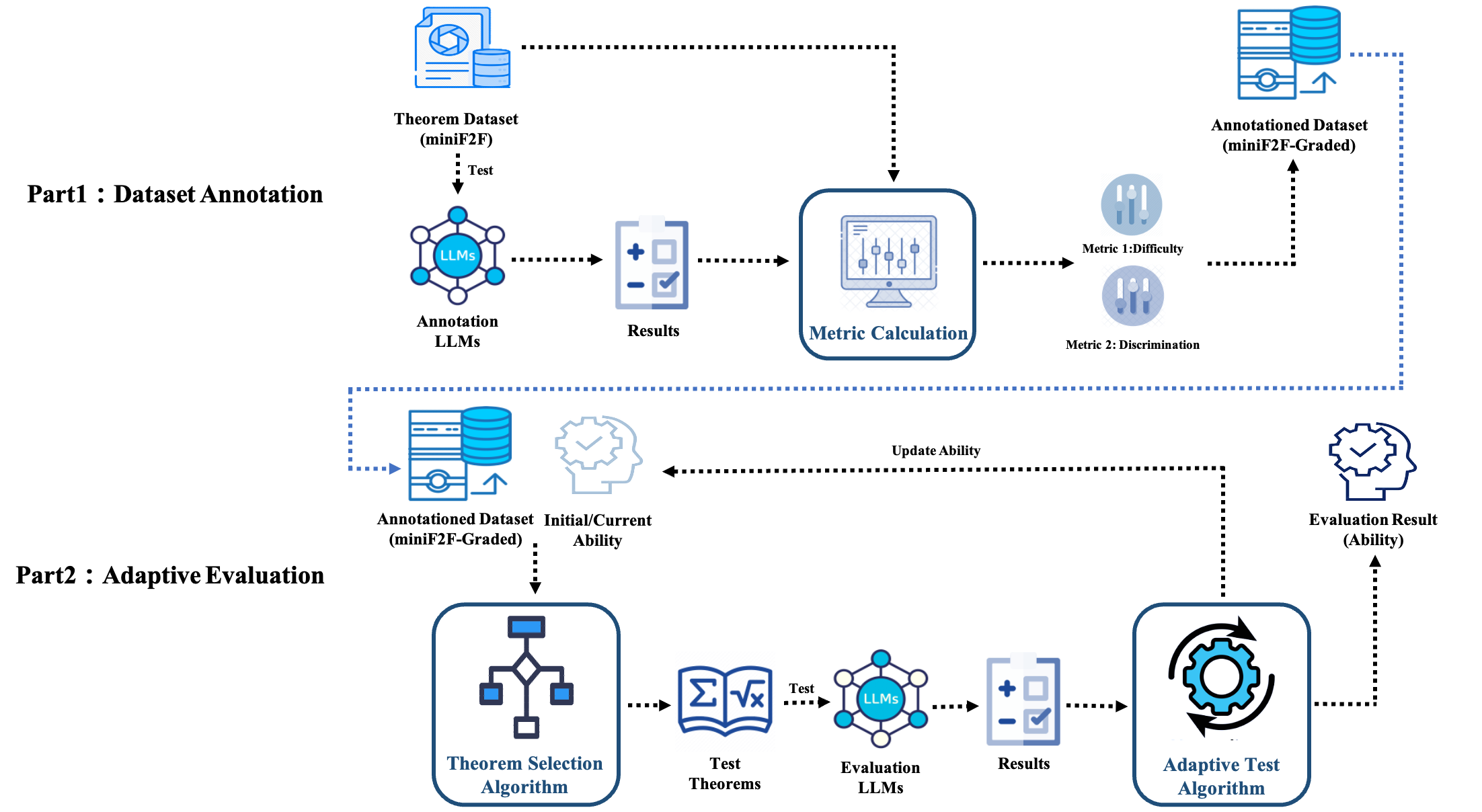}}
		\caption{The method consists of two parts. In Part 1: Dataset Annotation, difficulty and discrimination metrics are assigned to each theorem in the miniF2F dataset based on the performance of LLMs, as determined by the Metric Calculation process, resulting in an annotated dataset, miniF2F-Graded. In Part 2: Adaptive Evaluation, the most suitable theorems are dynamically selected for testing based on their annotated metrics in miniF2F-Graded and the initial or current ability scores of LLMs, following the Theorem Selection Algorithm. Throughout the iterative testing process, ability scores are continuously updated under the Adaptive Test Algorithm until they converge, with the final stabilized scores serving as the evaluation results.}
		\label{figure1}
	\end{center}
	\vskip -0.2in
\end{figure*}

To address these issues, this study draws on concepts from psychometrics: adaptive evaluation based on the model's real-time performance, rather than relying on a fixed test set. We propose a psychometric-based evaluation method for theorem proving with LLMs. This method annotates each theorem with difficulty (the ability level required for an LLM to prove it) and discrimination (how well the theorem distinguishes LLMs of varying abilities). Based on these annotations, the method dynamically selects the most suitable theorems to efficiently evaluate the LLM's proving ability. The overview of this method is shown in \ref{figure1}, consisting of two parts: Dataset Annotation (\ref{Dataset Annotation}) and Adaptive Evaluation (\ref{Adaptive Evaluation}).

In summary, the core contributions of this study are as follows: 
\begin{itemize} 
	
	\item This study proposes a psychometric-based evaluation method for theorem proving with LLMs, which evaluates the LLM’s proving ability through dataset annotation and adaptive evaluation, enabling a more fine-grained and efficient assessment. To the best of our knowledge, this is the first study to combine psychometrics with the evaluation of LLMs' theorem-proving ability. Specifically, this study designs a metric calculation method, a theorem selection algorithm, and an adaptive test algorithm, all tailored to the characteristics of LLMs in the context of theorem proving.

	\item Building on the widely used formal theorem dataset miniF2F, we propose an enhanced version with metric annotations and difficulty grading: miniF2F-Graded\footnote{The miniF2F-Graded dataset is available at \url{https://zenodo.org/records/14776138}.}. Experimental results indicate that, compared to the previous manual grading method, miniF2F-Graded more accurately reflects the difficulty of the theorems as perceived by LLMs. Additionally, this paper analyzes the distribution of metrics across the different categories of theorems in miniF2F, providing valuable guidance for future researchers in theorem proving, training, and evaluation.

	\item We apply our adaptive evaluation method and the miniF2F-Graded dataset to evaluate the performance of 10 open-source LLMs with theorem-proving abilities. The results demonstrate that our method finely distinguishes performance disparities between LLMs. Moreover, it uses only 23\% of the theorems in the dataset, thereby reducing the evaluation cost.
	
\end{itemize}
The structure of this paper is as follows:
\ref{section-2} reviews related work. \ref{section-3} introduces our method, including dataset annotation and adaptive evaluation. \ref{section-4} presents experiments verifying the effectiveness of our method. \ref{section-5} offers a detailed analysis of the experimental results. Finally, \ref{section-6} concludes the paper with a summary and directions for future work.

\section{Related Work}\label{section-2}
This study proposes a psychometric-based evaluation method for theorem proving with LLMs. Our research builds on existing literature in Theorem Proving with LLMs, LLM evaluation methods, and psychometrics.

\textbf{Theorem Proving with LLMs.} In recent years, a growing number of studies have explored training LLMs for theorem proving \cite{yang2024formal}. These works have developed models capable of generating proofs in formal languages such as Lean \cite{moura2021lean}, Isabelle \cite{paulson1994isabelle}, and Coq \cite{huet1997coq}. Based on proof generation strategies, existing methods can be broadly categorized into tree search methods and whole-proof generation methods. Tree search methods formulate the proof process as a search problem, incrementally exploring potential proof paths within a search tree. Notable approaches include LeanDojo \cite{yang2024leandojo}, Thor \cite{jiang2022thor}, LEGO \cite{wang2023lego}, DSP \cite{jiang2022draft}, among others. In contrast, whole-proof generation methods treat the proof as a complete sequence, generating the entire proof path in a single step. Representative works in this category include Baldur \cite{first2023baldur}, DeepSeek \cite{xin2024deepseekproverv15harnessingproofassistant}, TheoremLlama \cite{wang2024theoremllama}, and MetaMath-Llemma \cite{yu2023metamath}. Among them, DeepSeek-Prover-V1.5 leverages an enhanced formal theorem proving dataset for supervised fine-tuning, achieving a 60.2\% pass rate on the miniF2F test set using a single-pass full-proof generation approach \cite{xin2024deepseekproverv15harnessingproofassistant}.

In this study, we focus on evaluating whole-proof generation models, as their evaluation should adhere to consistent criteria. In contrast, tree search methods, due to their distinct search strategies, face challenges in establishing uniform evaluation standards.

\textbf{LLM Evaluation Methods.} Evaluating the capabilities of large language models remains a critical challenge. In the field of Theorem Proving with LLMs, proof pass rate is widely regarded as the primary evaluation criterion \cite{xia2024evaluating}. However, an increasing number of studies advocate for more comprehensive evaluation methods. For instance, \cite{zhang2025mathverse, hao2024llm} propose Chain-of-Thought (CoT) evaluation strategies, which assess LLM performance at each critical reasoning step rather than relying solely on the final answer. \cite{shao2024empirical, jin2023cladder, orenes2023using} highlight the importance of analyzing errors in the reasoning process to gain deeper insights into the common mistakes made by LLMs. \cite{xia2024evaluating, hong2024stuck} explore evaluation methods based on perturbation analysis, examining how LLMs respond to variations in input. Lastly, \cite{srivastava2024evaluating} suggest categorizing and annotating evaluation data based on the number of reasoning steps and question types, enabling a more detailed assessment of LLM performance across different data types.

\cite{zhuang2023efficiently} explores the use of human psychometrics in LLM evaluation and proposes that adaptive evaluation, which adjusts to a model’s performance rather than relying on fixed test sets, will become the new standard in AI model assessment. This concept serves as an important reference for our study. Specific applications of adaptive evaluation in LLM assessment can be found in works such as \cite{polo2024tinybenchmarks, yuan2024s}, which achieve efficient evaluation with fewer examples. This study aims to investigate how adaptive evaluation methods can be applied to evaluating the theorem-proving abilities of LLMs.

\textbf{Psychometrics.} Psychometrics is a field dedicated to the effective measurement of psychological traits. In exam assessments, it provides the scientific foundation and methodological tools for designing, analyzing, and interpreting exam results, ensuring that assessments accurately and reliably measure the true abilities of students or test subjects \cite{furr2021psychometrics, templin2010diagnostic, mislevy2003brief}. Item Response Theory (IRT), introduced by \cite{baker2001basics}, models both the characteristics of individual test items (e.g., difficulty, discrimination) and the ability scores of test subjects, offering a more precise measurement framework. IRT remains one of the most influential theoretical models in psychometrics \cite{fayers2004item}. Building on IRT’s approach to modeling item difficulty and discrimination, this study develops a metric calculation method and an adaptive test algorithm specifically designed to address the limited number of test subjects (LLMs) in theorem proving, achieving promising results.

\section{Method}\label{section-3}

\subsection{Dataset Annotation}\label{Dataset Annotation}

\subsubsection{miniF2F dataset}\label{3.2.1}
This study uses the MiniF2F dataset \cite{zheng2021minif2f} as the basis for annotation. The MiniF2F dataset is one of the most widely used datasets in the field, with many LLMs evaluating their performance based on the pass rate achieved on MiniF2F \cite{wang2023lego, xin2024deepseek}. A more detailed introduction of the MiniF2F dataset can be found in the \ref{appendix-Introduction to MiniF2F}.

Theorems in the MiniF2F dataset are formalized using a range of languages, including Metamath, Lean, Isabelle, and HOL Light. The difficulty of proving the same theorem can vary depending on the formal language used. For instance, a typical proof in Lean is significantly shorter than its counterpart in Metamath, as Lean offers many powerful tactics that aid in formalization \cite{zheng2021minif2f}. As a result, many studies opt to use Lean data for training and evaluation. To ensure the comparability of evaluation results, this study focuses on evaluating theorems in MiniF2F that are formalized in Lean. The dataset in \cite{yang23minif2f} updates the theorems from miniF2F to a newer version of Lean and includes natural language descriptions. Therefore, we use this version of miniF2F for annotation and evaluation.

In the MiniF2F paper \cite{zheng2021minif2f}, it is noted that 260 theorems in the dataset are derived from the MATH dataset \cite{hendrycks2021measuring}, with the theorems in MATH being classified into five difficulty levels. However, this grading was done manually by humans. We believe that the perceptions of difficulty between humans and LLMs may differ, as confirmed by the results in \ref{table3}. Therefore, in this study, we re-annotate each theorem's difficulty and discrimination, categorizing them into difficulty levels.

\subsubsection{Metric Calculation}\label{3.2.2}

This study argues that there is a similarity between LLM test sets and human exam questions: not all questions hold equal importance \cite{zhuang2023efficiently}. For instance, an LLM should receive higher scores for solving high-difficulty theorems than for solving low-difficulty ones. Furthermore, for an LLM that has already proven several relatively difficult theorems, testing a high-difficulty theorem is more valuable than testing a simpler one.

To fully exploit the differences between the theorems, this study uses the miniF2F dataset as an example and calculates the difficulty and discrimination of each theorem based on the performance of multiple LLMs (four Annotation LLMs were selected, as described in \ref{models-1}).

In Item Response Theory from psychometrics \cite{fayers2004item,cai2016item}, metrics like difficulty and discrimination are typically estimated through statistical models. A detailed explanation of the statistical modeling methods can be found in the \ref{appendix-Metric Calculation by Statistical Modeling}. By applying the statistical modeling methods from IRT and the theorem-proving results from LLMs, we estimated the difficulty and discrimination metrics for the theorems. However, the results were suboptimal, with the estimated metric distributions being too concentrated. The suboptimal results stem from the limited number of test subjects, as we used only four LLMs, far fewer than the number of examinees in human exams used in IRT models. However, it is currently impossible to find the same number of open-source LLMs as human test-takers.

To address the issue of insufficient test subjects, this study developed a metric calculation method that does not rely on metric estimation, instead using a more direct formula to compute the difficulty and discrimination metrics for the theorems.

\textbf{Notations and Definitions}
To accurately calculate the difficulty and discrimination metrics, we need to know the prior ability values of the models. We use the pass rate across the entire dataset ($Pass@128$) to represent the prior ability values of the models. The models used for annotation are denoted as $M_1, M_2, M_3, M_4$, with their relative prior ability values ranked as: ${\theta}(M_1)<{\theta}(M_2)<{\theta}(M_3)<{\theta}(M_4)$. We also compute the attempt success rate for each theorem across the four models. The attempt success rate is defined as the number of successful proof attempts divided by the total number of attempts, with each model making 128 attempts. The attempt success rates for the 4 models on theorem $x$ are represented as $P_{M_1}(x)$, $P_{M_2}(x)$, $P_{M_3}(x)$, and $P_{M_4}(x)$, and $P(x)$ is their average, representing the overall attempt success rate for theorem $x$. Finally, we use $Difficulty(x)$ and $Discrimination(x)$ to denote the difficulty and discrimination metrics for theorem $x$.

\textbf{Difficulty Calculation.}  
To calculate the difficulty more precisely, we designed a new formula that incorporates a correction term into $P(x)$, as shown in \ref{metric 1: Difficulty-1}. We subtract from $P(x)$ the value of each model that has successfully proven the theorem, divided by its prior ability score, and multiply the correction term by a weight $\epsilon$ (0.005). This approach aims to highlight the difficulty differences between the theorems. The more models that successfully prove the theorem, and the lower the ability of those models, the more it suggests that the theorem's difficulty should be lowered. We also experimented without the correction term, and the results are provided in the \ref{appendix-Difficulty Calculation with Omitting the Correction Term}. The findings indicate that including the correction term makes it easier to distinguish between the difficulty levels of the theorems.

\begin{equation}
	\label{metric 1: Difficulty-1}
	\begin{aligned}
		P^{\prime}(x) = P(x) - \epsilon \cdot \sum_{i=1}^{4} \mathbf{1}_{P_{M_i}(x) > 0} \cdot \frac{1}{\theta(M_{i})}.
	\end{aligned}
\end{equation}

We then use \ref{metric 1: Difficulty-2} to calculate the difficulty for each theorem. Instead of directly using the attempt success rate, we employ a reciprocal relationship derived from logistic regression. This approach more effectively captures the nonlinear relationship between the success rate and difficulty. The rationale behind this is that $P^{\prime}(x)$ exhibits smaller changes near 0 and 1, even though the actual difficulty may differ significantly. For example, the difference between $P^{\prime}(x)=0.01$ and $P^{\prime}(x)=0.1$ may be more substantial than the difference between $P^{\prime}(x)=0.6$ and $P^{\prime}(x)=0.7$. By amplifying the differences near 0 and 1, \ref{metric 1: Difficulty-2} highlights variations in the marginal intervals. This mapping enables a clearer distinction between high- and low-difficulty theorems by better separating items with low and high success rates.

\begin{equation}
	\label{metric 1: Difficulty-2}
	\begin{aligned}
		Difficulty(x)=-\frac{P^{\prime}(x)}{1-P^{\prime}(x)}.
	\end{aligned}
\end{equation}

\textbf{Discrimination Calculation.} 
We use \ref{metric 2: Discrimination-1} to calculate the discrimination for each theorem. $\mathcal{P}$ represents the set of all possible pairs of model combinations, which consists of 6 pairs.

\begin{equation}
	\label{metric 2: Discrimination-1}
	\begin{aligned}
		& Discrimination(x) = \frac{1}{6} \sum_{(i, j) \in \mathcal{P}} \frac{P_i(x) - P_j(x)}{\theta(i) - \theta(j)} \\
		& \text{where } \mathcal{P} = \{(i, j) \mid i, j \in \{M_1, M_2, M_3, M_4\}, i < j \}.
	\end{aligned}
\end{equation}

Using \ref{metric 2: Discrimination-1}, we calculate the ratio of the difference in attempt success rates to the difference in ability levels between two models for a given theorem. If the ability difference between the two models is small but the pass rate difference is large, it indicates that the theorem has high discrimination, effectively distinguishing between models of different ability levels. A negative discrimination value may suggest data contamination or that a low-ability model is particularly adept at proving the theorem, warranting further investigation. To reduce potential bias introduced by any single model, we sum the discrimination ratios for all model pairs and take the average, yielding a more stable discrimination metric.

To facilitate subsequent information content calculations, we linearly normalize the difficulty and discrimination metrics to the intervals $[0,1]$ and $[-1,1]$, respectively.

\subsection{Adaptive Evaluation}\label{Adaptive Evaluation}
\subsubsection{Theorem Selection Algorithm}  
After completing the dataset annotation, a theorem selection algorithm is developed to identify the most informative theorems for testing, based on their difficulty, discrimination, and the model's current ability score during adaptive evaluation. In Item Response Theory, the information of a question reflects how effectively it measures ability at a specific ability score \cite{cai2016item}. The greater the information, the better the question can differentiate between candidates with varying ability scores at that level, leading to more accurate measurements.

Information is typically calculated using the Fisher information function \cite{ly2017tutorial}. In this study, we apply a slightly modified version of the Fisher information function, as shown in \ref{selection-1}. Here, $P(x, \theta)$ represents the probability of theorem $x$ being correctly proven by a model with ability score $\theta$, calculated using the \ref{IRT-2pt} from Item Response Theory \cite{fayers2004item}. $a(x)$ denotes the discrimination, while $a(x)^f$ represents a modification introduced in this study to adjust and amplify the influence of the discrimination metric on the information, prioritizing the selection of theorems with higher discrimination. Here, $f$ is a hyperparameter.

\begin{equation}
	\label{selection-1}
	I(x,\theta) = a(x)^{f} \cdot P(x,\theta) \cdot (1 - P(x,\theta)).
\end{equation}

The selection algorithm is outlined in \ref{appendix-Theorem Selection Algorithm}. We rank the candidate theorems in $\mathcal{D}$ based on their information scores, excluding $\mathcal{T}{last}$, which have already been selected in the previous 10 rounds of testing to prevent overfitting the ability score by repeatedly choosing the same theorems. Finally, the top 5 theorems, $\mathcal{T}{selected}$, are selected for this round based on the information ranking.

\subsubsection{Adaptive Test Algorithm}
In the adaptive evaluation, our goal is to obtain an ability score that accurately reflects the model's theorem-proving ability. To achieve this, we perform multiple rounds of testing and ability score updates, gradually converging to the true ability score. As shown in \ref{test-1}, we first assign an initial ability score of 0.5. This initial score allows the model to begin testing with theorems of moderate difficulty. Next, based on the theorem selection algorithm, we select theorems for testing. The ability score, denoted as $\theta$, is updated according to the model's performance on these theorems, and new theorems are selected for further testing using the updated ability score. This process continues until the model's ability score changes by less than 0.01 over 10 consecutive rounds. At this point, we consider the model's ability score to have converged to its true value.

In \ref{test-1}, we designed the ability score update rule. When the success rate on a theorem $r_i$ is greater than 0 but less than 0.1, we consider that the model's ability to prove this theorem may be influenced by random factors. For instance, when the model’s success rate is 0.023, meaning it succeeded in only 3 out of 128 attempts, such a small number of successful attempts suggests a smaller score for this proof. Therefore, we apply a special transformation to these models: $r_i \gets \log(1 + r_i)$. The step size for updating the ability score is controlled by the hyperparameter $\eta$.

For the two hyperparameters appearing in the algorithm of this chapter: the discrimination influence factor $f$ in \ref{selection-1} and the step size adjustment factor $\eta$ in \ref{test-1}, we selected an optimal set of parameters through tuning experiments: ${[f, \eta]} = {[0.49, 0.004]}$. 

\begin{algorithm}[tb]
	\caption{Adaptive Test}
	\label{test-1}
	\begin{algorithmic}
		\STATE {\bfseries Input:} Model $M$, Annotated Dataset  $\mathcal{D}$ with metrics $Discrimination: a_i$ and $Difficulty: b_i$\\
		\STATE {\bfseries Output:} Ability Score $\theta$
		\STATE $\theta, \theta_{\text{prev}} \gets 0.5, S \gets 0$
		\WHILE{$S < $10}
		\STATE $\mathcal{T}_{selected} \gets \operatorname{\text{Theorem} \, \text{Selection}}(\theta, \mathcal{D})$ 
		\FOR{each theorem $t_i \in \mathcal{T}_{selected}$}
		\STATE $r_i \gets \operatorname{\text{Proof} \, \text{Test}(t_i, M)}$      /*Test and return the success rate of $M$ proving $t_i$ 128 times.*/
		\STATE \textbf{If} $0 < r_i < 0.1$ \textbf{Then} $r_i \gets \log(r_i + 1)$
		\ENDFOR
		\FOR{$i = 1$ \textbf{to} $|\mathcal{T}_{selected}|$}
		\STATE $P(t_i,\theta) \gets \frac{1}{1 + e^{-a_i (\theta - b_i)}}$
		\STATE $\theta \gets \theta + \eta \cdot a_i \cdot (r_i - P(t_i,\theta))$
		\ENDFOR
		\STATE $\theta \gets \min(\max(\theta, \theta_{\min}), \theta_{\max})$ 
		\STATE \textbf{Convergence Check:}
		\STATE \textbf{If} $|\theta - \theta_{\text{prev}}| < 0.01$ \textbf{Then} $S \gets S + 1$
		\STATE \textbf{Else} $S \gets 0$
		\STATE $\theta_{\text{prev}} \gets \theta$
		\ENDWHILE
		\STATE \textbf{Return} $\theta$
	\end{algorithmic}
\end{algorithm}

\section{Experiment}\label{section-4}

\subsection{Experiment Setup}
In this section, we introduce two distinct sets of LLMs: Annotation LLMs, used for dataset annotation, and Evaluation LLMs, responsible for adaptive evaluation, as described in \ref{section-3}. Additionally, we present the proof settings.

\textbf{Annotation LLMs.} \label{models-1}
When selecting the Annotation LLMs, we aimed to choose a model that could achieve the current state-of-the-art (SOTA) level for generating complete Lean proofs. To ensure more reasonable difficulty annotations for the theorems, we selected DeepSeek-Prover-V1.5-RL \cite{xin2024deepseekproverv15harnessingproofassistant} as one of the Annotation LLMs. This model achieved a $Pass@128$ of 58.61\% on miniF2F2 (including the validation and test sets; this result and the following pass rates were obtained through our own testing), which, to the best of our knowledge, represents the highest pass rate currently.

To achieve appropriate discrimination metrics, a sufficient gap in the proof capabilities of the Annotation LLMs is necessary. Therefore, we selected the following LLMs: codegeex4 \cite{zheng2023codegeex}, llemma \cite{azerbayev2023llemma}, TheoremLlama\cite{wang2024theoremllamatransforminggeneralpurposellms}, DeepSeek-Prover-V1.5-RL\cite{xin2024deepseekproverv15harnessingproofassistant}. The introduction to Annotation LLMs can be found in \ref{appendix-Annotation LLMs}.

\textbf{Evaluation LLMs.}  \label{models-2}
To validate whether our method can be applied to a broader range of models, we need to select a set of Evaluation LLMs distinct from the Annotation LLMs. Given the limited number of open-source models with theorem-proving capabilities, the selection process proved challenging. Nevertheless, we made efforts to identify models with theorem-proving ability as follows: Code-Llama \cite{roziere2023code}, Qwen2.5-Coder-7B \cite{qwen2,hui2024qwen2}, MetaMath-Llemma-7B \cite{yu2023metamath,azerbayev2023llemma}, DeepSeek-Prover-V1 \cite{xin2024deepseek}, DeepSeek-Prover-V1.5-Base. DeepSeek-Prover-V1.5-SFT \cite{xin2024deepseekproverv15harnessingproofassistant}. The introduction to Evaluation LLMs can be found in \ref{appendix-Evaluation LLMs}.

\textbf{Proof Settings.} 
When using the models mentioned above for proof generation, specific prompts must be provided. The DeepSeek-Prover series models and TheoremLlama were designed with prompts tailored for Lean proofs \cite{xin2024deepseekproverv15harnessingproofassistant, wang2024theoremllamatransforminggeneralpurposellms}, and we continued using these prompts in our testing. For other models, the prompts utilized are listed in \ref{appendix-Proof Prompt}.

We set the maximum number of proof attempts to 128, with a maximum proof length of 2048 tokens. The generated proofs are verified for correctness by interacting with Lean 4.13 \cite{moura2021lean}.

\subsection{Main Results}
\begin{figure}[ht]
	\vskip 0.2in
	\begin{center}
		\centerline{\includegraphics[width=\columnwidth]{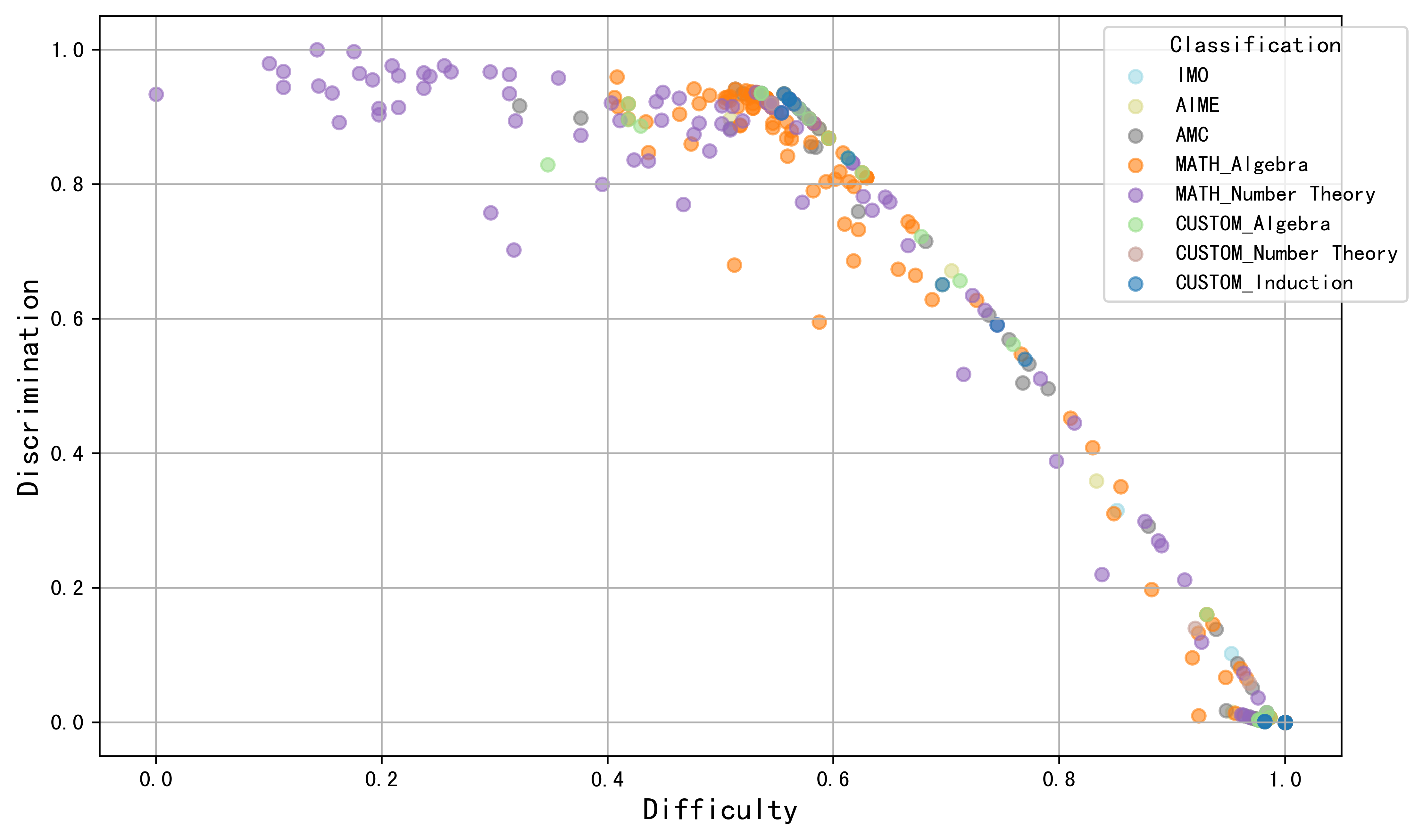}}
		\caption{The scatter plot of the dataset annotation results illustrates the relationship between theorem difficulty (x-axis) and discrimination (y-axis).}
		\label{figure2}
	\end{center}
	\vskip -0.2in
\end{figure}

\textbf{Dataset Annotation Results. }Using the annotation models described in \ref{models-1} and the methods outlined in \ref{Dataset Annotation}, we completed the annotation and grading of the dataset, resulting in the miniF2F-Graded dataset. The detailed results of the annotation and grading can be found in \ref{appendix-Data Annotation Results and Analysis}. We classified 488 theorems by type and created a scatter plot based on difficulty and discrimination, as shown in \ref{figure2}. It can be observed that, except for some theorems in mathd\_numbertheory, which exhibit high discrimination at lower difficulty levels due to one model's exceptional performance in this category, other theorems show an increase in discrimination as difficulty rises. The theorems within the difficulty range of 0.4–0.6 generally demonstrate higher discrimination, but as the difficulty continues to increase, the discrimination begins to decrease. This difficulty-discrimination distribution trend largely aligns with our understanding of test problems. For some unusual observations in the figure, we provide a detailed analysis in \ref{Analysis on Unusual Observations in the Scatter Plot}.

\textbf{Model Evaluation Results.} Based on the annotated dataset (miniF2F-Graded) and the adaptive evaluation method described in \ref{Adaptive Evaluation}, we evaluated six Evaluation LLMs and four Annotation LLMs. The evaluation results are presented in \ref{table1}. We also present in \ref{appendix-Evaluation Process} the process of model evaluation, including the selection of theorems, the changes in ability scores, and the evaluation results. It can be observed that the evaluation method proposed in this study significantly reduces the evaluation cost for both the Annotation LLMs and Evaluation LLMs, with an average reduction of 76.13\%. While the Ability Score and $Pass@128$ are measured in different units and have numerical differences, their rankings are largely the same. Further analysis of the authenticity of the evaluation results is discussed in \ref{Analysis on Model Evaluation Results}.

\begin{table}[ht]
	\caption{Model evaluation results. The table summarizes our evaluation outcomes: Ability Score. $Pass@128$ is included solely for comparison. The Evaluation Cost represents the number of theorems used in the evaluation, and Cost Changes indicates the reduction ratio of our method's Evaluation Cost relative to the full dataset.}
	\label{table1}
	\vskip 0.15in
	\begin{center}
		\resizebox{\textwidth}{!}{
			\begin{small}
				\begin{sc}
					\begin{tabular}{lcccc}
						\toprule
						\multicolumn{5}{c}{\textbf{Annotation LLMs}} \\
						\midrule
						Model & Ability Score & $Pass@128$ & Evaluation Cost & \textbf{Cost Changes} \\
						\midrule
						codegeex4 & \textbf{0.1038} & 0.0799 & 195 & \textbf{-60.04\%} \\
						llemma & \textbf{0.2946} & 0.2029 & 110 & \textbf{-77.46\%} \\
						TheoremLlama & \textbf{0.4283 }& 0.416 & 55 & \textbf{-88.73\%} \\
						DeepSeek-Prover-V1.5-RL & \textbf{0.6234} & 0.5861 & 65 & \textbf{-86.68\%} \\
						\midrule
						\multicolumn{5}{c}{\textbf{Evaluation LLMs}} \\
						\midrule
						Model & Ability Score & $Pass@128$ & Evaluation Cost & \textbf{Cost Changes} \\
						\midrule
						Code-Llama & \textbf{0.1384} & 0.123 & 180 & \textbf{-63.11\%} \\
						Qwen2.5-Coder & \textbf{0.2344 }& 0.1783 & 135 & \textbf{-72.34\%} \\
						MetaMath-Llemma &\textbf{ 0.2353} & 0.1967 & 135 & \textbf{-72.34\%} \\
						DeepSeek-Prover-V1.5-Base & \textbf{0.1427 }& 0.1803 & 180 & \textbf{-63.11\%} \\
						DeepSeek-Prover-V1 & \textbf{0.5323} & 0.5205 & 55 & \textbf{-88.73\%} \\
						DeepSeek-Prover-V1.5-SFT & \textbf{0.5995} & 0.582 & 55 & \textbf{-88.73\%} \\
						\bottomrule
					\end{tabular}
				\end{sc}
			\end{small}
		}
	\end{center}
	\vskip -0.1in
\end{table}

\section{Analysis on Results}\label{section-5}
\subsection{Analysis on Data Annotation Results}\label{section-5.1}

\textbf{Effectiveness of Difficulty Grading.}
We divided the miniF2F-graded theorems into Level 1-4 based on the sorting of the difficulty metric. Among them, Level 4 theorems are the most difficult, with a pass rate of 0 in all Annotation LLMs, corresponding to a difficulty metric of 1, and there are 127 theorems in this category. The remaining three levels are divided based on the principle of equal quantity, and the difficulty coefficient range and quantity for each level are shown in \ref{table2}.

\begin{table}[ht]
	\caption{Difficulty Grading.}
	\label{table2}
	\vskip 0.15in
	\begin{center}
		\resizebox{\textwidth}{!}{
			\begin{small}
			\begin{sc}
				\begin{tabular}{lccr}
					\toprule
					Difficulty Grading &  Difficulty range & Count \\
					\midrule
					Level 1    & $0 \leq x \leq 0.5539$ & 120 \\
					Level 2 & $0.5539 < x \leq 0.7661$ & 120 \\
					Level 3    & $0.7661 < x \leq 0.9864$ & 121 \\
					Level 4    & $x = 1$ & 127 \\
					\bottomrule
				\end{tabular}
			\end{sc}
		\end{small}
	}
	\end{center}
	\vskip -0.1in
\end{table}

To validate the effectiveness of the difficulty grading in miniF2F-graded, we tested the pass rates of various models on the theorems from the four difficulty levels and compared the results with the original grading method in the miniF2F dataset \cite{zheng2021minif2f}, as shown in \ref{table3}. It can be seen that in miniF2F-Graded, whether for all theorems or for MATH-category theorems, there is a clear distinction in the pass rates of LLMs across difficulty levels Level 1-4. For each LLM, the pass rate gradually decreases from Level 1 to Level 4, without exception. In contrast, in the original miniF2F, although the difficulty is divided into five levels, there is little distinction in the pass rates across different levels, and even the average pass rate for Level 2 theorems is higher than that for Level 1 theorems. The above results also confirm our hypothesis: LLMs perceive theorem difficulty differently from humans. miniF2F-Graded annotates difficulty based on the model's actual performance, which better reflects the difficulty of theorems as perceived by LLMs.

\begin{table}[ht]
	\caption{Model performance ($Pass@128$) under different difficulty levels by MiniF2F-Graded and MiniF2F. In the original miniF2F dataset, only 260 MATH-category theorems were manually graded into 5 difficulty levels, while the remaining theorems lack grading information. For comparison, we present separate results for all theorems and for only the MATH-category theorems in the table.}
	\label{table3}
	\vskip 0.15in
	\begin{center}
		\resizebox{\textwidth}{!}{
			\begin{small}
				\begin{sc}
					\begin{tabular}{lcccccc}
						\toprule
						\multicolumn{7}{c}{\textbf{miniF2F-Graded}} \\
						\midrule
						Model & Theorem Set & \textbf{Level 1 }&\textbf{ Level 2} & \textbf{Level 3} & \textbf{Level 4} \\
						\midrule
						Code-Llama-7b & Math & 0.48 & 0.0649 & 0.0217 & 0 \\
						& All & 0.45 & 0.0417 & 0.0083 & 0 \\
						Qwen2.5-Coder-7b & Math & 0.65 & 0.1558 & 0.0652 & 0 \\
						& All & 0.5917 & 0.1083 & 0.0248 & 0 \\
						MetaMath-Llemma-7b & Math & 0.68 & 0.2078 & 0.0652 & 0 \\
						& All & 0.6333 & 0.1417 & 0.0248 & 0 \\
						DeepSeek-Prover-V1.5-Base & Math & 0.61 & 0.2078 & 0.0217 & 0 \\
						& All & 0.5833 & 0.1333 & 0.0165 & 0 \\
						DeepSeek-Prover-V1 & Math & 1 & 0.9351 & 0.3696 & 0 \\
						& All & 1 & 0.925 & 0.1901 & 0 \\
						DeepSeek-Prover-V1.5-SFT & Math & 1 & 1 & 0.6087 & 0.027 \\
						& All & 1 & 1 & 0.3471 & 0.0157 \\
						\midrule
						\textbf{Average} & \textbf{Math} & \textbf{0.7367} & \textbf{0.4286} & \textbf{0.1920} & \textbf{0.0045} \\
						& \textbf{All} & \textbf{0.7097} & \textbf{0.3917} & \textbf{0.1019} & \textbf{0.0026} \\
						\midrule
						\multicolumn{7}{c}{\textbf{miniF2F}} \\
						\midrule
						Model & Theorem Set & \textbf{Level 1} & \textbf{Level 2} & \textbf{Level 3} & \textbf{Level 4} & \textbf{Level 5 }\\
						\midrule
						Code-Llama-7b & Math & 0.0952 & 0.2 & 0.26 & 0.2388 & 0.1806 \\
						Qwen2.5-Coder-7b & Math & 0.3333 & 0.44 & 0.28 & 0.2687 & 0.2639 \\
						MetaMath-Llemma-7b & Math & 0.381 & 0.42 & 0.3 & 0.3731 & 0.25 \\
						DeepSeek-Prover-V1.5-Base & Math & 0.4286 & 0.32 & 0.32 & 0.2687 & 0.2639 \\
						DeepSeek-Prover-V1 & Math & 0.7143 & 0.7 & 0.76 & 0.7015 & 0.75 \\
						DeepSeek-Prover-V1.5-SFT & Math & 0.8095 & 0.76 & 0.84 & 0.7463 & 0.8194 \\
						\midrule
						\textbf{Average} & \textbf{Math} & \textbf{0.4603} & \textbf{0.4733} & \textbf{0.46} & \textbf{0.43285} & \textbf{0.4213} \\
						\bottomrule
					\end{tabular}
				\end{sc}
			\end{small}
		}
	\end{center}
	\vskip -0.1in
\end{table}

\textbf{Metric Distribution of Theorems in Each Category.}
The theorems in miniF2F have multiple classification methods, and the detailed classification schemes can be found in the \ref{appendix-Theorem Categories} that introduces miniF2F. We have summarized the average difficulty and discrimination metrics for each category of theorems based on these classification methods, as shown in \ref{table4}. Additionally, we have plotted the distribution of difficulty metrics for theorems in each category and provided a detailed discussion of these distributions. This part can be found in \ref{appendix-Analysis on Distribution of Difficulty Metrics}.

\begin{table}[ht]
	\caption{The average values of difficulty and discrimination metrics for theorems in each classification under different categorization methods.}
	\label{table4}
	\vskip 0.15in
	\begin{center}
		\resizebox{\textwidth}{!}{\begin{small}
				\begin{tabular}{lcccc}
					\toprule
					Categorization & Classification & Difficulty & Discrimination & Count \\
					\midrule
					Overall & / & 0.7494 & 0.4493 & 488 \\
					\midrule
					{Sort by Usage} 
					& Test & 0.8118 & 0.3243 & 244 \\
					& Validation & 0.687 & 0.5743 & 244 \\
					\midrule
					{Sort by Provenance} 
					& IMO & 0.978 & 0.0335 & 40 \\
					& AIME & 0.9177 & 0.1486 & 30 \\
					& AMC & 0.8571 & 0.2774 & 90 \\
					& MATH & 0.6458 & 0.6185 & 260 \\
					& CUSTOM & 0.794 & 0.4072 & 68 \\
					\midrule
					{Sort by Problem} 
					& Algebra & 0.7012 & 0.5787 & 136 \\
					& Number Theory & 0.6318 & 0.5838 & 176 \\
					& Induction & 0.7854 & 0.4523 & 16 \\
					& Others & 0.8987 & 0.1923 & 160 \\
					\bottomrule
				\end{tabular}
			\end{small}
		}
	\end{center}
	\vskip -0.1in
\end{table}

Based on the data in \ref{table4} and the analysis in Appendix, we can conclude that the metric distribution of theorems in each category aligns with our expectations. For instance, when categorized by source, it is clear that the difficulty values of theorems from math competitions such as IMO, AIME, and AMC are significantly higher than those from the CUSTOM (mathematics courses) category. Several examples analyzed in the \ref{appendix-Analysis on Distribution of Difficulty Metrics} further support the reasonableness of our metric calculation.

\subsection{Analysis on  Model Evaluation Results}\label{Analysis on Model Evaluation Results}
\begin{figure}[ht]
	\vskip 0.2in
	\begin{center}
		\centerline{\includegraphics[width=\columnwidth]{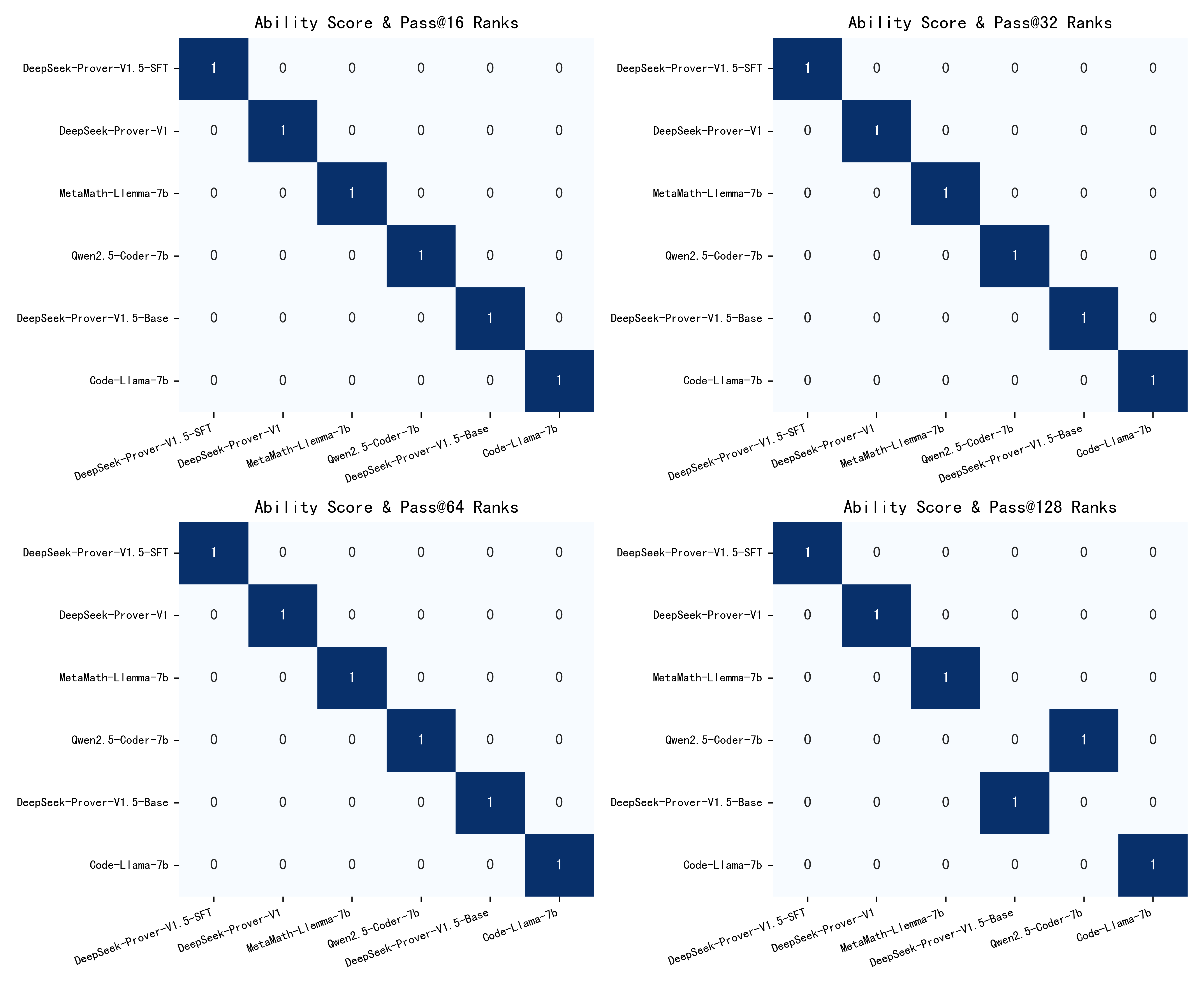}}
		\caption{Confusion matrix of model rankings.}
		\label{figure3}
	\end{center}
	\vskip -0.2in
\end{figure}

\textbf{Authenticity of Evaluation Results.}
In \ref{table1}, we present the evaluation results obtained using our method. This section examines whether these results accurately reflect the true proof abilities of the LLMs. The results from our evaluation method are represented by ability scores, which belong to a different dimension than pass rates and cannot be directly compared. However, we can still compare the differences in proof abilities between models as reflected by these two metrics.We compared the pass rate rankings of the LLMs with different numbers of attempts to the rankings obtained from our ability scores, and the resulting confusion matrix is shown in \ref{figure3}. From the results in the figure, it is clear that the ability score rankings align well with those of $Pass@16$, $Pass@32$, and $Pass@64$. The only exception is in the comparison with $Pass@128$, where one pair of models—Qwen2.5-Coder-7B and DeepSeek-Prover-V1.5-Base—are ranked differently. We provide a detailed analysis of this group of models with differing rankings in \ref{appendix-Analysis on Ability Ranking}. The \ref{appendix-Analysis on Ability Ranking} also discusses why our results better highlight the ability differences across multiple sets of models, such as DeepSeek-Prover-V1.5-RL and DeepSeek-Prover-V1.5-SFT.

In conclusion, the ability scores obtained through this evaluation method accurately reflect the proof abilities of LLMs. Compared to pass rates, ability scores better reveal the performance disparities among LLMs.

\section{Conclusion and Future Work}\label{section-6}
This study proposes a psychometric-based evaluation method for theorem proving with LLMs, which includes dataset annotation and adaptive evaluation. It effectively leverages the varying importance of theorems, enabling accurate evaluation of proof abilities with a minimal number of high-information theorems. By annotating the miniF2F dataset, this research introduces a new dataset with metric annotations and difficulty grading: miniF2F-Graded. The difficulty grading in miniF2F-Graded is more closely aligned with LLMs' difficulty perception compared to manual grading. In the evaluation of multiple open-source LLMs, the adaptive evaluation method reduced the evaluation cost by 76.13\% and yielded results that more clearly highlight the performance disparities between LLMs.

Currently, this method has been implemented using the miniF2F dataset as a case study. However, we believe the approach is applicable to more extensive and diverse datasets and is expected to offer even greater efficiency improvements when applied to larger-scale data collections. In the future, we plan to expand the method to additional datasets and models, continuously updating the annotations and grading in miniF2F-Graded using progressively enhanced SOTA models, thereby providing a reliable and efficient evaluation benchmark for LLMs in theorem proving.

\section*{Impact Statement}

This paper presents a psychometric-based evaluation method for large language models in theorem proving. By introducing the miniF2F-Graded dataset, which includes difficulty and discrimination metrics. The adaptive evaluation method efficiently highlights performance disparities between LLMs using fewer theorems.

The societal impact of this research lies in its potential to optimize LLM evaluation in automated theorem proving, making the process more efficient and reliable. Ethical considerations focus on ensuring the fairness and transparency of the grading system. Future work will extend this method to larger datasets and models, providing a scalable benchmark for LLM evaluation in theorem proving.


\bibliographystyle{unsrt}  
\bibliography{references}

\begin{thebibliography}{10}

\bibitem{naveed2023comprehensive}
Humza Naveed, Asad~Ullah Khan, Shi Qiu, Muhammad Saqib, Saeed Anwar, Muhammad
  Usman, Naveed Akhtar, Nick Barnes, and Ajmal Mian.
\newblock A comprehensive overview of large language models.
\newblock {\em arXiv preprint arXiv:2307.06435}, 2023.

\bibitem{zhao2023survey}
Wayne~Xin Zhao, Kun Zhou, Junyi Li, Tianyi Tang, Xiaolei Wang, Yupeng Hou,
  Yingqian Min, Beichen Zhang, Junjie Zhang, Zican Dong, et~al.
\newblock A survey of large language models.
\newblock {\em arXiv preprint arXiv:2303.18223}, 2023.

\bibitem{yang2024formal}
Kaiyu Yang, Gabriel Poesia, Jingxuan He, Wenda Li, Kristin Lauter, Swarat
  Chaudhuri, and Dawn Song.
\newblock Formal mathematical reasoning: A new frontier in ai.
\newblock {\em arXiv preprint arXiv:2412.16075}, 2024.

\bibitem{ahn2024large}
Janice Ahn, Rishu Verma, Renze Lou, Di~Liu, Rui Zhang, and Wenpeng Yin.
\newblock Large language models for mathematical reasoning: Progresses and
  challenges.
\newblock {\em arXiv preprint arXiv:2402.00157}, 2024.

\bibitem{xin2024deepseekproverv15harnessingproofassistant}
Huajian Xin, Z.~Z. Ren, Junxiao Song, Zhihong Shao, Wanjia Zhao, Haocheng Wang,
  Bo~Liu, Liyue Zhang, Xuan Lu, Qiushi Du, Wenjun Gao, Qihao Zhu, Dejian Yang,
  Zhibin Gou, Z.~F. Wu, Fuli Luo, and Chong Ruan.
\newblock Deepseek-prover-v1.5: Harnessing proof assistant feedback for
  reinforcement learning and monte-carlo tree search.
\newblock 2024.

\bibitem{wang2023lego}
Haiming Wang, Huajian Xin, Chuanyang Zheng, Lin Li, Zhengying Liu, Qingxing
  Cao, Yinya Huang, Jing Xiong, Han Shi, Enze Xie, et~al.
\newblock Lego-prover: Neural theorem proving with growing libraries.
\newblock {\em arXiv preprint arXiv:2310.00656}, 2023.

\bibitem{xia2024evaluating}
Shijie Xia, Xuefeng Li, Yixin Liu, Tongshuang Wu, and Pengfei Liu.
\newblock Evaluating mathematical reasoning beyond accuracy.
\newblock {\em arXiv preprint arXiv:2404.05692}, 2024.

\bibitem{moura2021lean}
Leonardo~de Moura and Sebastian Ullrich.
\newblock The lean 4 theorem prover and programming language.
\newblock In {\em Automated Deduction--CADE 28: 28th International Conference
  on Automated Deduction, Virtual Event, July 12--15, 2021, Proceedings 28},
  pages 625--635. Springer, 2021.

\bibitem{paulson1994isabelle}
Lawrence~C Paulson.
\newblock {\em Isabelle: A generic theorem prover}.
\newblock Springer, 1994.

\bibitem{huet1997coq}
G{\'e}rard Huet, Gilles Kahn, and Christine Paulin-Mohring.
\newblock The coq proof assistant a tutorial.
\newblock {\em Rapport Technique}, 178, 1997.

\bibitem{yang2024leandojo}
Kaiyu Yang, Aidan Swope, Alex Gu, Rahul Chalamala, Peiyang Song, Shixing Yu,
  Saad Godil, Ryan~J Prenger, and Animashree Anandkumar.
\newblock Leandojo: Theorem proving with retrieval-augmented language models.
\newblock {\em Advances in Neural Information Processing Systems}, 36, 2024.

\bibitem{jiang2022thor}
Albert~Qiaochu Jiang, Wenda Li, Szymon Tworkowski, Konrad Czechowski, Tomasz
  Odrzyg{\'o}{\'z}d{\'z}, Piotr Mi{\l}o{\'s}, Yuhuai Wu, and Mateja Jamnik.
\newblock Thor: Wielding hammers to integrate language models and automated
  theorem provers.
\newblock {\em Advances in Neural Information Processing Systems},
  35:8360--8373, 2022.

\bibitem{jiang2022draft}
Albert~Q Jiang, Sean Welleck, Jin~Peng Zhou, Wenda Li, Jiacheng Liu, Mateja
  Jamnik, Timoth{\'e}e Lacroix, Yuhuai Wu, and Guillaume Lample.
\newblock Draft, sketch, and prove: Guiding formal theorem provers with
  informal proofs.
\newblock {\em arXiv preprint arXiv:2210.12283}, 2022.

\bibitem{first2023baldur}
Emily First, Markus~N Rabe, Talia Ringer, and Yuriy Brun.
\newblock Baldur: Whole-proof generation and repair with large language models.
\newblock In {\em Proceedings of the 31st ACM Joint European Software
  Engineering Conference and Symposium on the Foundations of Software
  Engineering}, pages 1229--1241, 2023.

\bibitem{wang2024theoremllama}
Ruida Wang, Jipeng Zhang, Yizhen Jia, Rui Pan, Shizhe Diao, Renjie Pi, and Tong
  Zhang.
\newblock Theoremllama: Transforming general-purpose llms into lean4 experts.
\newblock {\em arXiv preprint arXiv:2407.03203}, 2024.

\bibitem{yu2023metamath}
Longhui Yu, Weisen Jiang, Han Shi, Jincheng Yu, Zhengying Liu, Yu~Zhang,
  James~T Kwok, Zhenguo Li, Adrian Weller, and Weiyang Liu.
\newblock Metamath: Bootstrap your own mathematical questions for large
  language models.
\newblock {\em arXiv preprint arXiv:2309.12284}, 2023.

\bibitem{zhang2025mathverse}
Renrui Zhang, Dongzhi Jiang, Yichi Zhang, Haokun Lin, Ziyu Guo, Pengshuo Qiu,
  Aojun Zhou, Pan Lu, Kai-Wei Chang, Yu~Qiao, et~al.
\newblock Mathverse: Does your multi-modal llm truly see the diagrams in visual
  math problems?
\newblock In {\em European Conference on Computer Vision}, pages 169--186.
  Springer, 2025.

\bibitem{hao2024llm}
Shibo Hao, Yi~Gu, Haotian Luo, Tianyang Liu, Xiyan Shao, Xinyuan Wang, Shuhua
  Xie, Haodi Ma, Adithya Samavedhi, Qiyue Gao, et~al.
\newblock Llm reasoners: New evaluation, library, and analysis of step-by-step
  reasoning with large language models.
\newblock {\em arXiv preprint arXiv:2404.05221}, 2024.

\bibitem{shao2024empirical}
Minghao Shao, Boyuan Chen, Sofija Jancheska, Brendan Dolan-Gavitt, Siddharth
  Garg, Ramesh Karri, and Muhammad Shafique.
\newblock An empirical evaluation of llms for solving offensive security
  challenges.
\newblock {\em arXiv preprint arXiv:2402.11814}, 2024.

\bibitem{jin2023cladder}
Zhijing Jin, Yuen Chen, Felix Leeb, Luigi Gresele, Ojasv Kamal, LYU Zhiheng,
  Kevin Blin, Fernando~Gonzalez Adauto, Max Kleiman-Weiner, Mrinmaya Sachan,
  et~al.
\newblock Cladder: Assessing causal reasoning in language models.
\newblock In {\em Thirty-seventh conference on neural information processing
  systems}, 2023.

\bibitem{orenes2023using}
Marcelo Orenes-Vera, Margaret Martonosi, and David Wentzlaff.
\newblock Using llms to facilitate formal verification of rtl.
\newblock {\em arXiv e-prints}, pages arXiv--2309, 2023.

\bibitem{hong2024stuck}
Pengfei Hong, Deepanway Ghosal, Navonil Majumder, Somak Aditya, Rada Mihalcea,
  and Soujanya Poria.
\newblock Stuck in the quicksand of numeracy, far from agi summit: Evaluating
  llms' mathematical competency through ontology-guided perturbations.
\newblock {\em arXiv preprint arXiv:2401.09395}, 2024.

\bibitem{srivastava2024evaluating}
Pragya Srivastava, Manuj Malik, Vivek Gupta, Tanuja Ganu, and Dan Roth.
\newblock Evaluating llms’ mathematical reasoning in financial document
  question answering.
\newblock In {\em Findings of the Association for Computational Linguistics ACL
  2024}, pages 3853--3878, 2024.

\bibitem{zhuang2023efficiently}
Yan Zhuang, Qi~Liu, Yuting Ning, Weizhe Huang, Rui Lv, Zhenya Huang, Guanhao
  Zhao, Zheng Zhang, Qingyang Mao, Shijin Wang, et~al.
\newblock Efficiently measuring the cognitive ability of llms: An adaptive
  testing perspective.
\newblock {\em arXiv preprint arXiv:2306.10512}, 2023.

\bibitem{polo2024tinybenchmarks}
Felipe~Maia Polo, Lucas Weber, Leshem Choshen, Yuekai Sun, Gongjun Xu, and
  Mikhail Yurochkin.
\newblock tinybenchmarks: evaluating llms with fewer examples.
\newblock {\em arXiv preprint arXiv:2402.14992}, 2024.

\bibitem{yuan2024s}
Xiaohan Yuan, Jinfeng Li, Dongxia Wang, Yuefeng Chen, Xiaofeng Mao, Longtao
  Huang, Hui Xue, Wenhai Wang, Kui Ren, and Jingyi Wang.
\newblock S-eval: Automatic and adaptive test generation for benchmarking
  safety evaluation of large language models.
\newblock {\em arXiv preprint arXiv:2405.14191}, 2024.

\bibitem{furr2021psychometrics}
R~Michael Furr.
\newblock {\em Psychometrics: an introduction}.
\newblock SAGE publications, 2021.

\bibitem{templin2010diagnostic}
Jonathan Templin, Robert~A Henson, et~al.
\newblock {\em Diagnostic measurement: Theory, methods, and applications}.
\newblock Guilford press, 2010.

\bibitem{mislevy2003brief}
Robert~J Mislevy, Russell~G Almond, and Janice~F Lukas.
\newblock A brief introduction to evidence-centered design.
\newblock {\em ETS Research Report Series}, 2003(1):i--29, 2003.

\bibitem{baker2001basics}
Frank~B Baker.
\newblock {\em The basics of item response theory}.
\newblock ERIC, 2001.

\bibitem{fayers2004item}
Peter Fayers.
\newblock Item response theory for psychologists, 2004.

\bibitem{zheng2021minif2f}
Kunhao Zheng, Jesse~Michael Han, and Stanislas Polu.
\newblock Minif2f: a cross-system benchmark for formal olympiad-level
  mathematics.
\newblock {\em arXiv preprint arXiv:2109.00110}, 2021.

\bibitem{xin2024deepseek}
Huajian Xin, Daya Guo, Zhihong Shao, Zhizhou Ren, Qihao Zhu, Bo~Liu, Chong
  Ruan, Wenda Li, and Xiaodan Liang.
\newblock Deepseek-prover: Advancing theorem proving in llms through
  large-scale synthetic data.
\newblock {\em arXiv preprint arXiv:2405.14333}, 2024.

\bibitem{yang23minif2f}
K.~Yang.
\newblock minif2f-lean4, 2023.

\bibitem{hendrycks2021measuring}
Dan Hendrycks, Collin Burns, Saurav Kadavath, Akul Arora, Steven Basart, Eric
  Tang, Dawn Song, and Jacob Steinhardt.
\newblock Measuring mathematical problem solving with the math dataset.
\newblock {\em arXiv preprint arXiv:2103.03874}, 2021.

\bibitem{cai2016item}
Li~Cai, Kilchan Choi, Mark Hansen, and Lauren Harrell.
\newblock Item response theory.
\newblock {\em Annual Review of Statistics and Its Application}, 3(1):297--321,
  2016.

\bibitem{ly2017tutorial}
Alexander Ly, Maarten Marsman, Josine Verhagen, Raoul~PPP Grasman, and Eric-Jan
  Wagenmakers.
\newblock A tutorial on fisher information.
\newblock {\em Journal of Mathematical Psychology}, 80:40--55, 2017.

\bibitem{zheng2023codegeex}
Qinkai Zheng, Xiao Xia, Xu~Zou, Yuxiao Dong, Shan Wang, Yufei Xue, Zihan Wang,
  Lei Shen, Andi Wang, Yang Li, Teng Su, Zhilin Yang, and Jie Tang.
\newblock Codegeex: A pre-trained model for code generation with multilingual
  benchmarking on humaneval-x.
\newblock In {\em Proceedings of the 29th ACM SIGKDD Conference on Knowledge
  Discovery and Data Mining}, pages 5673--5684, 2023.

\bibitem{azerbayev2023llemma}
Zhangir Azerbayev, Hailey Schoelkopf, Keiran Paster, Marco~Dos Santos, Stephen
  McAleer, Albert~Q. Jiang, Jia Deng, Stella Biderman, and Sean Welleck.
\newblock Llemma: An open language model for mathematics, 2023.

\bibitem{wang2024theoremllamatransforminggeneralpurposellms}
Ruida Wang, Jipeng Zhang, Yizhen Jia, Rui Pan, Shizhe Diao, Renjie Pi, and Tong
  Zhang.
\newblock Theoremllama: Transforming general-purpose llms into lean4 experts,
  2024.

\bibitem{roziere2023code}
Baptiste Roziere, Jonas Gehring, Fabian Gloeckle, Sten Sootla, Itai Gat,
  Xiaoqing~Ellen Tan, Yossi Adi, Jingyu Liu, Romain Sauvestre, Tal Remez,
  et~al.
\newblock Code llama: Open foundation models for code.
\newblock {\em arXiv preprint arXiv:2308.12950}, 2023.

\bibitem{qwen2}
An~Yang, Baosong Yang, Binyuan Hui, Bo~Zheng, Bowen Yu, Chang Zhou, Chengpeng
  Li, Chengyuan Li, Dayiheng Liu, Fei Huang, Guanting Dong, Haoran Wei, Huan
  Lin, Jialong Tang, Jialin Wang, Jian Yang, Jianhong Tu, Jianwei Zhang,
  Jianxin Ma, Jin Xu, Jingren Zhou, Jinze Bai, Jinzheng He, Junyang Lin, Kai
  Dang, Keming Lu, Keqin Chen, Kexin Yang, Mei Li, Mingfeng Xue, Na~Ni, Pei
  Zhang, Peng Wang, Ru~Peng, Rui Men, Ruize Gao, Runji Lin, Shijie Wang, Shuai
  Bai, Sinan Tan, Tianhang Zhu, Tianhao Li, Tianyu Liu, Wenbin Ge, Xiaodong
  Deng, Xiaohuan Zhou, Xingzhang Ren, Xinyu Zhang, Xipin Wei, Xuancheng Ren,
  Yang Fan, Yang Yao, Yichang Zhang, Yu~Wan, Yunfei Chu, Yuqiong Liu, Zeyu Cui,
  Zhenru Zhang, and Zhihao Fan.
\newblock Qwen2 technical report.
\newblock {\em arXiv preprint arXiv:2407.10671}, 2024.

\bibitem{hui2024qwen2}
Binyuan Hui, Jian Yang, Zeyu Cui, Jiaxi Yang, Dayiheng Liu, Lei Zhang, Tianyu
  Liu, Jiajun Zhang, Bowen Yu, Kai Dang, et~al.
\newblock Qwen2. 5-coder technical report.
\newblock {\em arXiv preprint arXiv:2409.12186}, 2024.

\bibitem{jiang2021lisa}
Albert~Qiaochu Jiang, Wenda Li, Jesse~Michael Han, and Yuhuai Wu.
\newblock Lisa: Language models of isabelle proofs.
\newblock In {\em 6th Conference on Artificial Intelligence and Theorem
  Proving}, pages 378--392, 2021.

\bibitem{azerbayev2023proofnet}
Zhangir Azerbayev, Bartosz Piotrowski, Hailey Schoelkopf, Edward~W Ayers,
  Dragomir Radev, and Jeremy Avigad.
\newblock Proofnet: Autoformalizing and formally proving undergraduate-level
  mathematics.
\newblock {\em arXiv preprint arXiv:2302.12433}, 2023.

\bibitem{deepseek-math}
Qihao Zhu Runxin Xu Junxiao Song Mingchuan Zhang Y.K. Li Y. Wu Daya~Guo
  Zhihong~Shao, Peiyi~Wang.
\newblock Deepseekmath: Pushing the limits of mathematical reasoning in open
  language models, 2024.

\end{thebibliography}

\appendix
\onecolumn
\section{MiniF2F Dataset and Theorem Categories.}\label{appendix-MiniF2F Dataset and Theorem Categories}
\subsection{Introduction to MiniF2F}\label{appendix-Introduction to MiniF2F}
The miniF2F dataset \cite{zheng2021minif2f} comprises formalized mathematical problem statements at the Olympiad level. It includes 488 formalized problems spanning high school and college-level mathematics, as well as questions from prestigious competitions such as the International Mathematical Olympiad (IMO), American Mathematical Competitions (AMC), and American Invitational Mathematics Examination (AIME). Although smaller in size compared to datasets like LISA \cite{jiang2021lisa} and ProofNet \cite{azerbayev2023proofnet}, miniF2F is widely recognized for its high-quality data and extensive adoption, making it an ideal benchmark dataset for this study.

\subsection{Theorem Categories}\label{appendix-Theorem Categories}
The theorems in miniF2F are classified using multiple criteria. By purpose, they are divided into a test set and a validation set. By source, they are categorized into IMO (International Mathematical Olympiad), AIME (American Invitational Mathematics Examination), AMC (American Mathematics Competitions), MATH \cite{hendrycks2021measuring}, and CUSTOM (datasets from high school and undergraduate mathematics courses). By problem type, they are classified into number theory, algebra, and induction.

\section{Different Methods of Metric Calculation.}\label{appendix-Experiments with Different Methods of Metric Calculation}
\subsection{Metric Calculation by Statistical Modeling.}\label{appendix-Metric Calculation by Statistical Modeling}
In Item Response Theory (IRT), models are categorized based on the number of metrics they incorporate, such as one-metric, two-metric, or three-metric models. The two-metric model considers difficulty, which represents the ability score required to correctly answer a question, and discrimination, which measures how effectively a question differentiates between varying ability levels \cite{cai2016item}. The mathematical formulation for the two-metric model is as follows:

\begin{equation}
	\label{IRT-2pt}
	P(X=1 \mid \theta,x)=\frac{e^{a(x)(\theta-b(x))}}{1+e^{a(x)(\theta-b(x))}}.
\end{equation}

In \ref{IRT-2pt}, $\theta$ represents the ability level of the test subject, $a(x)$ and $b(x)$ represent the discrimination and difficulty of item x, respectively, and $P(X=1 \mid \theta, x)$ represents the probability of correctly answering item x with an ability level of $\theta$.

After determining the statistical model, methods such as Maximum Likelihood Estimation (MLE), Bayesian Estimation, and the EM algorithm are commonly used to estimate the difficulty and discrimination metrics. The more data from test-takers, the more accurate the estimates become.

\subsection{Difficulty Calculation without the Correction Term.}\label{appendix-Difficulty Calculation with Omitting the Correction Term}
\begin{figure}[ht]
	\vskip 0.2in
	\begin{center}
		\centerline{\includegraphics[width=\columnwidth]{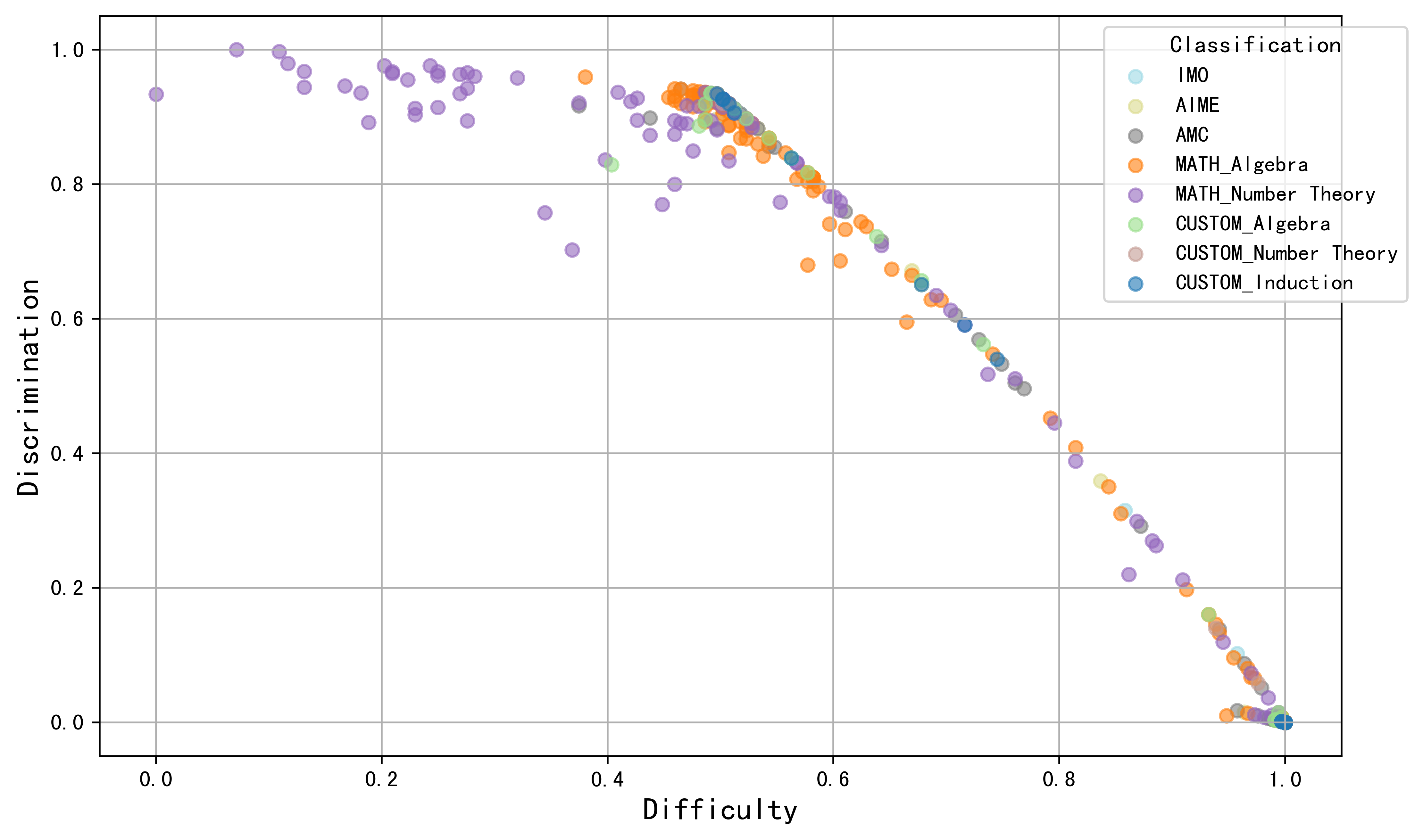}}
		\caption{The scatter plot of the dataset annotation results when the correction term is not used in the difficulty metric calculation.}
		\label{scatter_by_classification_without_correction}
	\end{center}
	\vskip -0.2in
\end{figure}

In the calculation of the difficulty metric, if we do not add a correction term to $P(x)$, the result of the metric calculation is shown in \ref{scatter_by_classification_without_correction}. By comparing \ref{scatter_by_classification_without_correction} and \ref{figure2}, we can observe that adding the correction term to the difficulty metric calculation results in a more dispersed distribution of difficulty values. Specifically, for the theorems with difficulty values between 0.4 and 0.6, the points in \ref{scatter_by_classification_without_correction} are almost clustered together, making it difficult to distinguish them. However, in \ref{figure2}, where the correction term is included, these theorems are spread across multiple difficulty levels, which is more conducive to theorem grading and selection.

\section{Theorem Selection Algorithm.}\label{appendix-Theorem Selection Algorithm}
\begin{algorithm}[H]
	\caption{Theorem Selection}
	\label{selection-2}
	\begin{algorithmic}
		\STATE {\bfseries Input:} Current ability value $\theta$,
		Annotated Dataset $\mathcal{D}$ with metrics $Discrimination: a_i$ and $Difficulty: b_i$
		\STATE {\bfseries Output:} Selected theorems $\mathcal{T}_{selected}$
		\IF{First execution}
		\STATE Initialize $\mathcal{T}_{last} = \emptyset$
		\ENDIF
		\FOR{each theorem $t_i \in \mathcal{D}$}
		\STATE $P(t_i,\theta) \gets \frac{1}{1 + e^{-a_i (\theta - b_i)}}$
		\STATE $I(t_i,\theta) \gets a_i^{f} \cdot P(t_i,\theta) \cdot (1 - P(t_i,\theta))$
		\ENDFOR
		\STATE $\mathcal{D}_{filtered} \gets \mathcal{D} \setminus \mathcal{T}_{last}$
		\STATE Initialize $\mathcal{T}_{selected} = \emptyset$
		\FOR{$k = 1$ {\bfseries to} $5$}
		\STATE $t^* \gets \arg \max_{t_i \in \mathcal{D}_{filtered}} I(t_i,\theta)$
		\STATE $\mathcal{T}_{selected} \gets \mathcal{T}_{selected} \cup \{t^*\}$
		\STATE $\mathcal{D}_{filtered} \gets \mathcal{D}_{filtered} \setminus \{t^*\}$
		\ENDFOR
		\STATE $\mathcal{T}_{last} \gets (\mathcal{T}_{last} \cup \mathcal{T}_{selected})[-50:]$
		\STATE \textbf{Return} $\mathcal{T}_{selected}$
	\end{algorithmic}
\end{algorithm}

\section{Experiment LLMs and Prompts.}\label{appendix-Experiment LLMs and Prompts}
\subsection{Annotation LLMs.}\label{appendix-Annotation LLMs}
1)codegeex4-9B (Pass@128=7.99\%) is the open-source version of the latest CodeGeeX4 model series. It is a multilingual code generation model trained on GLM-4-9B, which significantly enhances its code generation capabilities\cite{zheng2023codegeex}.

2)llemma-7B (Pass@128=20.29\%) is a mathematical language model. The Llemma model is particularly outstanding in terms of chain-of-thought mathematical reasoning and formal theorem proving\cite{azerbayev2023llemma}.

3)TheoremLlama (Pass@128=41.60\%) is a fine-tuned model for writing Lean4 proofs. Using the Open Bootstrapped Theorem dataset for fine-tuning, TheoremLlama can achieve quite good performance when writing Lean4 proofs based on natural language theorem statements\cite{wang2024theoremllamatransforminggeneralpurposellms}.

4)DeepSeek-Prover-V1.5-RL (Pass@128=58.61\%) is an open-source language model for theorem proving in Lean 4. The model was pre-trained on DeepSeekMath-Base, focusing on formal mathematical language, and uses reinforcement learning (RL) to improve performance based on the validation feedback of the Lean4 prover, resulting in a significant increase in proving ability\cite{xin2024deepseekproverv15harnessingproofassistant}.

\subsection{Evaluation LLMs.}\label{appendix-Evaluation LLMs}
1)Code-Llama-7B (Pass@128=12.30\%) is a large code language model series based on Llama 2, which has achieved state-of-the-art performance among open models in multiple code benchmarks\cite{roziere2023code}.

2)Qwen2.5-Coder-7B (Pass@128=17.83\%) is the latest series of the Code-Specific Qwen large language model. This model has expanded the training token scale to 5.5 trillion on the basis of Qwen2.5, not only enhancing its coding capabilities but also maintaining its advantages in mathematics and general abilities\cite{qwen2,hui2024qwen2}.

3)MetaMath-Llemma-7B (Pass@128=19.67\%), based on the Llemma-7B model, has been fully fine-tuned on the MetaMathQA dataset. The MetaMath-7B model has achieved 66.4\% on GSM8K and 19.4\% on MATH, surpassing the most advanced models of the same size\cite{yu2023metamath,azerbayev2023llemma}.

4-6) Three different versions of the DeepSeek-Prover series: DeepSeek-Prover-V1 (Pass@128=52.05\%), V1.5-Base (Pass@128=18.03\%), and V1.5-SFT (Pass@128=58.20\%). V1 fine-tuned the DeepSeekMath-7B model using large-scale synthetic data to enhance its proving capability\cite{xin2024deepseek}; V1.5-Base was pre-trained on DeepSeekMath-Base\cite{deepseek-math}, with a special focus on formal languages such as Lean, Isabelle, and Metamath. V1.5-SFT was supervised fine-tuned using an enhanced formal theorem proving dataset derived from DeepSeek-Prover-V1, which included detailed explanatory annotations.\cite{xin2024deepseekproverv15harnessingproofassistant}.

The DeepSeek-Prover series models are used multiple times in both the annotation and evaluation models for the following reasons: By comparing the evaluation results between different versions of the DeepSeek-Prover series models, we can verify the effectiveness of the evaluation methods used in this study. For example, DeepSeek-Prover-V1.5-RL, compared to DeepSeek-Prover-V1.5-SFT, has enhanced its capabilities through a combination of online reinforcement learning (RL) and validation feedback. However, the difference between the two in terms of Pass@128 is only 0.41\% \cite{xin2024deepseekproverv15harnessingproofassistant}. We aim to assess a more accurate gap between the two using the methods of this study.

Additionally, among the aforementioned models, some are not specifically trained for Lean formal proof but are targeted at tasks such as code generation and mathematics. However, judging by the pass rates, these models possess a certain level of formal proof capability, and their training data should include corpora written in Lean. Therefore, we believe they can be used for annotation and evaluation.

\subsection{Proof Prompt.}\label{appendix-Proof Prompt}
We used the following prompt to instruct the LLMs to perform theorem proofs, this is an example of a prompt for proving a Lean theorem. The prompt includes the natural language description of the theorem (Informal Prefix) from \cite{yang23minif2f}.

\begin{itemize} 
	\item \textbf{Prompt:} This is a theorem written in Lean 4. Please complete the corresponding proof code to formalize the argument:	
	-Informal Prefix:
	Expand the following expression: $7(3y+2)$ Show that it is $21y+14$.	
	-Header:
	import Mathlib
	import Aesop
	set\_option maxHeartbeats 0
	open BigOperators Real Nat Topology Rat
	-Formal Statement:
	$(y : \mathbb{C}) : 7 \cdot (3 \cdot y + 2) = 21 \cdot y + 14 \colon= \text{by}$
\end{itemize}

\section{Data Annotation Results and Analysis.}\label{appendix-Data Annotation Results and Analysis}
\subsection{Annotation Results: miniF2F-Graded.}\label{appendix-Annotation Results: miniF2F-Graded}
The first 60 theorems in our annotated results are shown in \ref{theorem-table}, The complete results can be found at \url{https://zenodo.org/records/14776138}.
\begin{table}[H]
	\caption{The annotation results of the first 60 theorems in miniF2F-Graded.}
	\label{theorem-table}
	\vskip 0.15in
	\begin{center}
		\begin{small}
			\begin{sc}
				\begin{tabular}{lcccc}
					\toprule
					Theorem Name & Difficulty & Discrimination & Difficulty Grade \\
					\midrule
					amc12a\_2019\_p21 & 1.0000 & 0.0000 & Level 4 \\
					amc12a\_2015\_p10 & 1.0000 & 0.0000 & Level 4 \\
					amc12a\_2008\_p8 & 0.7377 & 0.6056 & Level 2 \\
					mathd\_algebra\_182 & 0.4083 & 0.9594 & Level 1 \\
					aime\_1984\_p5 & 0.9793 & 0.0025 & Level 3 \\
					mathd\_numbertheory\_780 & 0.9819 & 0.0012 & Level 3 \\
					mathd\_algebra\_116 & 0.5083 & 0.9305 & Level 1 \\
					mathd\_numbertheory\_13 & 1.0000 & 0.0000 & Level 4 \\
					mathd\_numbertheory\_169 & 0.5012 & 0.8897 & Level 1 \\
					amc12a\_2009\_p9 & 0.5561 & 0.9339 & Level 2 \\
					amc12a\_2019\_p9 & 1.0000 & 0.0000 & Level 4 \\
					mathd\_algebra\_13 & 0.5405 & 0.9218 & Level 1 \\
					induction\_sum2kp1npqsqm1 & 0.6128 & 0.8390 & Level 2 \\
					aime\_1991\_p6 & 0.9767 & 0.0037 & Level 3 \\
					mathd\_numbertheory\_149 & 0.4107 & 0.8946 & Level 1 \\
					imo\_1984\_p2 & 1.0000 & 0.0000 & Level 4 \\
					amc12a\_2008\_p4 & 0.9682 & 0.0085 & Level 3 \\
					imo\_2006\_p3 & 1.0000 & 0.0000 & Level 4 \\
					mathd\_algebra\_462 & 0.4182 & 0.9193 & Level 1 \\
					imo\_1964\_p1\_2 & 0.9524 & 0.1021 & Level 3 \\
					mathd\_numbertheory\_221 & 1.0000 & 0.0000 & Level 4 \\
					mathd\_numbertheory\_64 & 0.5606 & 0.9266 & Level 2 \\
					imo\_1987\_p4 & 0.9819 & 0.0012 & Level 3 \\
					mathd\_numbertheory\_33 & 0.5450 & 0.9205 & Level 1 \\
					amc12\_2001\_p9 & 0.5651 & 0.9193 & Level 2 \\
					imo\_1965\_p1 & 1.0000 & 0.0000 & Level 4 \\
					mathd\_numbertheory\_48 & 0.5360 & 0.9351 & Level 1 \\
					numbertheory\_sqmod4in01d & 0.5561 & 0.9339 & Level 2 \\
					mathd\_numbertheory\_466 & 0.4430 & 0.9225 & Level 1 \\
					mathd\_algebra\_48 & 0.5174 & 0.8870 & Level 1 \\
					amc12\_2000\_p15 & 1.0000 & 0.0000 & Level 4 \\
					mathd\_numbertheory\_132 & 0.1974 & 0.9030 & Level 1 \\
					amc12a\_2009\_p5 & 0.5561 & 0.9339 & Level 2 \\
					mathd\_numbertheory\_188 & 0.2371 & 0.9427 & Level 1 \\
					mathd\_algebra\_224 & 1.0000 & 0.0000 & Level 4 \\
					induction\_divisibility\_3divnto3m2n & 0.5561 & 0.9339 & Level 2 \\
					induction\_sum\_1oktkp1 & 0.5606 & 0.9266 & Level 2 \\
					mathd\_numbertheory\_32 & 0.5539 & 0.9059 & Level 1 \\
					mathd\_algebra\_422 & 0.5827 & 0.8901 & Level 2 \\
					mathd\_algebra\_73 & 0.5268 & 0.9316 & Level 1 \\
					mathd\_numbertheory\_109 & 0.4634 & 0.9279 & Level 1 \\
					algebra\_xmysqpymzsqpzmxsqeqxyz\_xpypzp6dvdx3y3z3 & 0.5561 & 0.9339 & Level 2 \\
					amc12b\_2002\_p11 & 1.0000 & 0.0000 & Level 4 \\
					imo\_1962\_p4 & 1.0000 & 0.0000 & Level 4 \\
					mathd\_numbertheory\_236 & 0.2146 & 0.9611 & Level 1 \\
					mathd\_numbertheory\_24 & 0.4903 & 0.8495 & Level 1 \\
					algebra\_amgm\_prod1toneq1\_sum1tongeqn & 0.9819 & 0.0012 & Level 3 \\
					mathd\_algebra\_101 & 0.5606 & 0.9266 & Level 2 \\
					mathd\_numbertheory\_257 & 0.5651 & 0.9193 & Level 2 \\
					...... & ...... & ...... & ...... \\
					\bottomrule
				\end{tabular}
			\end{sc}
		\end{small}
	\end{center}
	\vskip -0.1in
\end{table}

\subsection{Analysis on Unusual Observations in the Scatter Plot.}\label{Analysis on Unusual Observations in the Scatter Plot}

The difficulty-discrimination distribution in \ref{figure2} generally aligns with our intuitive understanding of test items. However, unlike typical test distributions, no tester (LLM) has been able to correctly answer all difficult questions (theorems), leading to a blank upper-right region in the scatter plot. Additionally, the lower-right corner contains numerous points (where difficulty is close to 1 and discrimination is close to 0), representing theorems that no model has successfully proven. Furthermore, the absence of theorems with a discrimination value below 0 suggests that the proof performance of the selected labeling models exhibits a gradient distribution, which aligns with the prior ability distribution.

\subsection{Analysis on Distribution of Difficulty Metrics.}\label{appendix-Analysis on Distribution of Difficulty Metrics}
\begin{figure}[H]
	\vskip 0.2in
	\begin{center}
		\centerline{\includegraphics[width=\columnwidth]{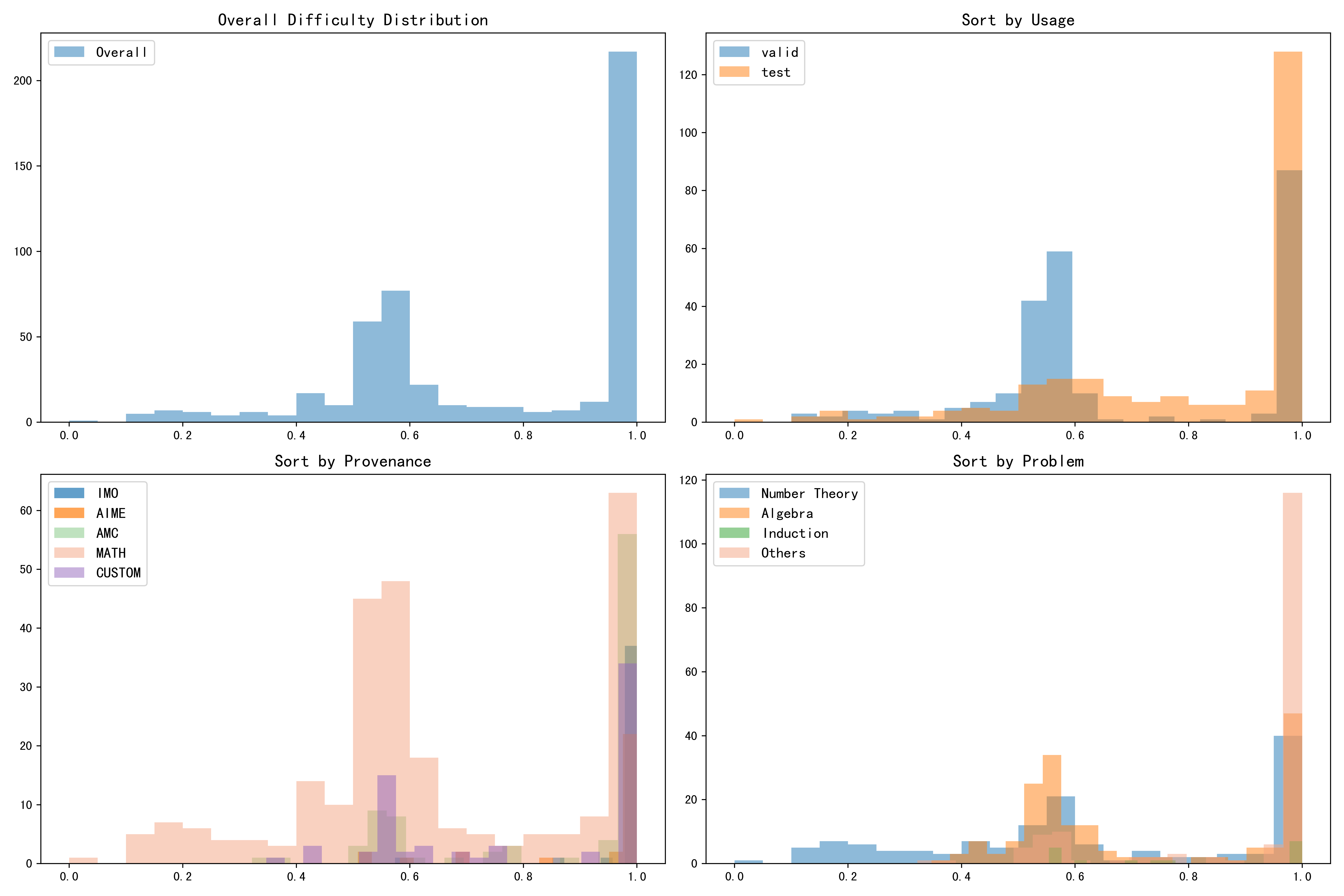}}
		\caption{Difficulty Distribution Chart for Theorems in Various Categories..}
		\label{difficulty_distribution_analysis}
	\end{center}
	\vskip -0.2in
\end{figure}

From the distribution in \ref{difficulty_distribution_analysis} and \ref{table4}, we can draw several key conclusions:

1) For all the theorems, excluding those with a difficulty value of 1, the difficulty of the remaining theorems roughly follows a normal distribution centered around 0.6.

2) When the theorems are divided into the test set and the validation set, it is evident that the overall difficulty of the test set is higher than that of the validation set, which aligns with our expectations. This is because, in the miniF2F design, the test set is reserved for evaluation, while the validation set may have been used during model training \cite{zheng2021minif2f}. However, the average discrimination of the validation set is still higher than that of the test set, which is due to the fact that the test set contains many high-difficulty theorems that reduce the discrimination.

3) When the theorems are categorized by source, it is clear that the average difficulty of the IMO, AIME, and AMC competitions is significantly higher than that of MATH and CUSTOM. This is because competition problems tend to be more complex, but their higher complexity also leads to a lower discrimination. The MATH dataset, with its moderate difficulty, has the highest discrimination, making it suitable for testing. In contrast, problems in the CUSTOM category, sourced from high school and undergraduate courses, have the lowest difficulty, which lowers the discrimination since even lower-capacity models can solve these problems. 

4) When classified by problem type, we can observe that induction problems are the most difficult for the models, while number theory problems are the easiest. Theorems in the "OTHERS" category, most of which originate from competitions, may involve algebra, number theory, and other types of problems, thus making their difficulty the highest.

\section{Adaptive Evaluation Results and Analysis.}\label{appendix-Adaptive Evaluation Results and Analysis}
\subsection{Evaluation Process.}\label{appendix-Evaluation Process}
We present the evaluation process of one LLM in \ref{test-results-table}.
\begin{table}[H]
	\caption{Test process for model DeepSeek-Prover-V1.}
	\label{test-results-table}
	\vskip 0.15in
	\begin{center}
		\resizebox{0.85\textwidth}{!}{
			\begin{small}
				\begin{sc}
					\begin{tabular}{lcccccc}
						\toprule
						Number & Theorem Name & Difficulty & Discrimination & Success Rate & Ability Score \\
						\midrule
						1 & mathd\_algebra\_182 & 0.4083 & 0.9594 & 0.5859375 & 0.50025 (+0.00025) \\
						2 & mathd\_numbertheory\_551 & 0.3564 & 0.9578 & 0.65625 & 0.50074 (+0.00049) \\
						3 & mathd\_numbertheory\_517 & 0.2552 & 0.9762 & 0.8671875 & 0.50197 (+0.00123) \\
						4 & mathd\_numbertheory\_198 & 0.2959 & 0.9668 & 0.7265625 & 0.50268 (+0.00071) \\
						5 & mathd\_numbertheory\_101 & 0.3128 & 0.9631 & 0.8046875 & 0.50372 (+0.00104) \\
						6 & mathd\_numbertheory\_202 & 0.1751 & 0.9971 & 0.890625 & 0.50496 (+0.00124) \\
						7 & mathd\_algebra\_109 & 0.4763 & 0.9419 & 0.78125 & 0.50605 (+0.00109) \\
						8 & mathd\_algebra\_536 & 0.513 & 0.9414 & 0.3046875 & 0.50528 (-0.00077) \\
						9 & amc12a\_2016\_p3 & 0.513 & 0.9414 & 0.484375 & 0.50522 (-0.00006) \\
						10 & mathd\_numbertheory\_37 & 0.2611 & 0.9669 & 0.6328125 & 0.50552 (+0.00030) \\
						11 & mathd\_algebra\_77 & 0.5223 & 0.9389 & 0.0234375 & 0.50363 (-0.00189) \\
						12 & mathd\_algebra\_55 & 0.5268 & 0.9376 & 0.9296875 & 0.50537 (+0.00174) \\
						13 & mathd\_algebra\_126 & 0.5314 & 0.9364 & 0.8125 & 0.50664 (+0.00127) \\
						14 & mathd\_algebra\_410 & 0.5314 & 0.9364 & 0.7578125 & 0.50770 (+0.00106) \\
						15 & mathd\_numbertheory\_412 & 0.5314 & 0.9364 & 0.796875 & 0.50891 (+0.00121) \\
						16 & mathd\_algebra\_182 & 0.4083 & 0.9594 & 0.5859375 & 0.50915 (+0.00024) \\
						17 & mathd\_numbertheory\_551 & 0.3564 & 0.9578 & 0.65625 & 0.50963 (+0.00048) \\
						18 & mathd\_numbertheory\_517 & 0.2552 & 0.9762 & 0.8671875 & 0.51085 (+0.00122) \\
						19 & mathd\_numbertheory\_198 & 0.2959 & 0.9668 & 0.7265625 & 0.51155 (+0.00070) \\
						20 & mathd\_numbertheory\_101 & 0.3128 & 0.9631 & 0.8046875 & 0.51258 (+0.00103) \\
						21 & mathd\_algebra\_536 & 0.513 & 0.9414 & 0.3046875 & 0.51180 (-0.00078) \\
						22 & amc12a\_2016\_p3 & 0.513 & 0.9414 & 0.484375 & 0.51174 (-0.00006) \\
						23 & mathd\_numbertheory\_202 & 0.1751 & 0.9971 & 0.890625 & 0.51297 (+0.00123) \\
						24 & mathd\_algebra\_109 & 0.4763 & 0.9419 & 0.78125 & 0.51406 (+0.00109) \\
						25 & mathd\_numbertheory\_37 & 0.2611 & 0.9669 & 0.6328125 & 0.51435 (+0.00029) \\
						26 & mathd\_algebra\_77 & 0.5223 & 0.9389 & 0.0234375 & 0.51245 (-0.00190) \\
						27 & mathd\_algebra\_55 & 0.5268 & 0.9376 & 0.9296875 & 0.51418 (+0.00173) \\
						28 & mathd\_algebra\_126 & 0.5314 & 0.9364 & 0.8125 & 0.51545 (+0.00127) \\
						29 & mathd\_algebra\_410 & 0.5314 & 0.9364 & 0.7578125 & 0.51649 (+0.00104) \\
						30 & mathd\_numbertheory\_412 & 0.5314 & 0.9364 & 0.796875 & 0.51769 (+0.00120) \\
						31 & mathd\_algebra\_182 & 0.4083 & 0.9594 & 0.5859375 & 0.51793 (+0.00024) \\
						32 & mathd\_numbertheory\_551 & 0.3564 & 0.9578 & 0.65625 & 0.51840 (+0.00047) \\
						33 & mathd\_numbertheory\_198 & 0.2959 & 0.9668 & 0.7265625 & 0.51910 (+0.00070) \\
						34 & mathd\_numbertheory\_101 & 0.3128 & 0.9631 & 0.8046875 & 0.52012 (+0.00102) \\
						35 & mathd\_numbertheory\_517 & 0.2552 & 0.9762 & 0.8671875 & 0.52133 (+0.00121) \\
						36 & mathd\_algebra\_536 & 0.513 & 0.9414 & 0.3046875 & 0.52054 (-0.00079) \\
						37 & amc12a\_2016\_p3 & 0.513 & 0.9414 & 0.484375 & 0.52047 (-0.00007) \\
						38 & mathd\_algebra\_109 & 0.4763 & 0.9419 & 0.78125 & 0.52155 (+0.00108) \\
						39 & mathd\_numbertheory\_202 & 0.1751 & 0.9971 & 0.890625 & 0.52277 (+0.00122) \\
						40 & mathd\_numbertheory\_37 & 0.2611 & 0.9669 & 0.6328125 & 0.52305 (+0.00028) \\
						41 & mathd\_algebra\_77 & 0.5223 & 0.9389 & 0.0234375 & 0.52114 (-0.00191) \\
						42 & mathd\_algebra\_55 & 0.5268 & 0.9376 & 0.9296875 & 0.52287 (+0.00173) \\
						43 & mathd\_algebra\_126 & 0.5314 & 0.9364 & 0.8125 & 0.52413 (+0.00126) \\
						44 & mathd\_algebra\_410 & 0.5314 & 0.9364 & 0.7578125 & 0.52516 (+0.00103) \\
						45 & mathd\_numbertheory\_412 & 0.5314 & 0.9364 & 0.796875 & 0.52636 (+0.00120) \\
						46 & mathd\_algebra\_182 & 0.4083 & 0.9594 & 0.5859375 & 0.52659 (+0.00023) \\
						47 & mathd\_numbertheory\_551 & 0.3564 & 0.9578 & 0.65625 & 0.52705 (+0.00046) \\
						48 & mathd\_numbertheory\_198 & 0.2959 & 0.9668 & 0.7265625 & 0.52773 (+0.00068) \\
						49 & mathd\_numbertheory\_101 & 0.3128 & 0.9631 & 0.8046875 & 0.52875 (+0.00102) \\
						50 & mathd\_numbertheory\_517 & 0.2552 & 0.9762 & 0.8671875 & 0.52995 (+0.00120) \\
						51 & mathd\_algebra\_536 & 0.513 & 0.9414 & 0.3046875 & 0.52915 (-0.00080) \\
						52 & amc12a\_2016\_p3 & 0.513 & 0.9414 & 0.484375 & 0.52908 (-0.00007) \\
						53 & mathd\_algebra\_109 & 0.4763 & 0.9419 & 0.78125 & 0.53015 (+0.00107) \\
						54 & mathd\_numbertheory\_202 & 0.1751 & 0.9971 & 0.890625 & 0.53136 (+0.00121) \\
						55 & mathd\_numbertheory\_48 & 0.536 & 0.9351 & 0.7421875 & 0.53234 (+0.00098) \\
						\bottomrule
					\end{tabular}
				\end{sc}
			\end{small}
		}
	\end{center}
	\vskip -0.1in
\end{table}

%

\subsection{Analysis on Ability Ranking}\label{appendix-Analysis on Ability Ranking}
We will analyze this pair of reversed models: In the ability score results, Qwen2.5-Coder-7B has a higher ability score (0.2344) than DeepSeek-Prover-V1.5-Base (0.1427), but in the $Pass@128$ results, the ranking is reversed. This occurs because Qwen2.5-Coder-7B performs better on higher-difficulty problems than DeepSeek-Prover-V1.5-Base. From \ref{table3}, we observe that Qwen2.5-Coder-7B achieves a pass rate of 6.52\% on Level 3 theorems, much higher than the 2.17\% achieved by DeepSeek-Prover-V1.5-Base. Our adaptive evaluation method assigns higher weights to high-difficulty theorems, which results in Qwen2.5-Coder-7B having a higher overall ability score. Furthermore, although DeepSeek-Prover-V1.5-Base (0.1803) slightly exceeds Qwen2.5-Coder-7B (0.1783) in $Pass@128$, when we compare the attempt success rate shown in \ref{table1-appendix}, Qwen2.5-Coder-7B (0.0068) is higher than DeepSeek-Prover-V1.5-Base (0.0055). This suggests that the higher ability score assigned to Qwen2.5-Coder-7B by our evaluation method is reasonable.

Analysis of other models:
Taking the codegeex4-9b model as an example, the model’s $Pass@128$ reaches 0.0799, but the attempt success rate is only 0.0012, which is much lower than Llemma-7b’s 0.0119. This indicates that, out of 128 proof attempts, the model has only a small number of successful attempts. The $Pass@128$ metric clearly ignores this aspect, focusing only on whether there was a success in any of the 128 attempts. The actual proof capability gap between codegeex4-9b and Llemma-7b should be greater than what is indicated by $Pass@128$. Our method’s evaluation result, 0.1038, accounts for the number of successful attempts and provides a more authentic reflection of the actual proof capability gap, which is more accurately shown by Llemma-7b (0.2946).

Additionally, comparing DeepSeek-Prover-V1.5-RL and SFT, the RL model is enhanced with online reinforcement learning and other techniques based on the SFT model. However, the improvement in the $Pass@128$ metric is only 0.41\%. Is the actual difference between the two models really that small? We can see from the attempt success rate that this is not the case. The attempt success of RL (47.72\%) is higher than that of SFT (44.86\%), and our evaluation method has more accurately reflected the difference between them, with ability scores of 0.623 and 0.5995, respectively.

\begin{table*}[ht]
	\caption{Comparison of Evaluation Results.}
	\label{table1-appendix}
	\vskip 0.15in
	\begin{center}
		\resizebox{\textwidth}{!}{
		\begin{scriptsize}
			\begin{tabular}{lccc}
				\toprule
				\textbf{Model} & 
				\textbf{Ability Score} & 
				\textbf{Pass@128} & 
				\textbf{Attempt Succse Rate} \\
				\midrule
				Codegeex4-9b & \textbf{0.1038} & 0.0799 & 0.0012 \\
				Llemma\-7b & \textbf{0.2946} & 0.2029 & 0.0119 \\
				TheoremLlama & \textbf{0.4283} & 0.416 & 0.026 \\
				DeepSeek-Prover-V1.5-RL & \textbf{0.6234} & 0.5861 & 0.4772 \\
				Code-Llama-7b & \textbf{0.1384} & 0.123 & 0.0022 \\
				Qwen2.5-Coder-7b & \textbf{0.2344} & 0.1783 & 0.0068 \\
				MetaMath-Llemma-7b & \textbf{0.2353} & 0.1967 & 0.0094 \\
				DeepSeek-Prover-V1.5-Base & \textbf{0.1427} & 0.1803 & 0.0055 \\
				DeepSeek-Prover-V1 & \textbf{0.5323} & 0.5205 & 0.3282 \\
				DeepSeek-Prover-V1.5-SFT & \textbf{0.5995} & 0.5820 & 0.4486 \\
				\bottomrule
			\end{tabular}
		\end{scriptsize}
	}
	\end{center}
	\vskip -0.1in
\end{table*}

\end{document}